\documentclass{article}

 \usepackage[main, final]{neurips_2026}

\usepackage[utf8]{inputenc} 
\usepackage[T1]{fontenc}    
\usepackage{hyperref}       
\usepackage{url}            
\usepackage{booktabs}       
\usepackage{amsfonts}       
\usepackage{nicefrac}       
\usepackage{microtype}      
\usepackage{xcolor}         

\usepackage{graphicx}
\usepackage[table]{xcolor}
\usepackage{amsmath}
\usepackage{amssymb}
\usepackage{mathtools}
\usepackage{amsthm}
\usepackage{enumitem}
\usepackage{dsfont}
\usepackage{tabularx}
\usepackage{ragged2e} 
\usepackage{multirow}
\usepackage{caption}
\usepackage{algorithm}
\usepackage{algpseudocode}

\newcommand{\red}[1]{\textcolor{red}{#1}}
\newcommand{\blue}[1]{\textcolor{blue}{#1}}

\title{Revisiting the Generic Transformer: Deconstructing a Strong Baseline for Time Series Foundation Models}

\author{%
  Yunshi Wen\thanks{Work done while the author was a Ph.D. student at Rensselaer Polytechnic Institute.\protect\newline\hspace*{1.8em}Correspondence to: \texttt{yunshi-wen@outlook.com}.}\\
  Rensselaer Polytechnic \\
  Institute\\
  \And
  Wesley M. Gifford \\
  IBM Research \\
  \texttt{wmgifford@us.ibm.com} \\
  \And
  Chandra Reddy \\
  IBM Research \\
  \texttt{creddy@us.ibm.com} \\
  \And
  Lam M. Nguyen \\
  IBM Research \\
  \texttt{lamnguyen.mltd@ibm.com} \\
  \And
  Jayant Kalagnanam \\
  IBM Research \\
  \texttt{jayant@us.ibm.com} \\
  \And
  Anak Agung Julius \\
  Rensselaer Polytechnic \\
  Institute \\
  \texttt{agung@ecse.rpi.edu} \\
}

\begin{document}

\maketitle
\setcounter{footnote}{0}

\begin{abstract}
The recent surge of interest in Time Series Foundation Models has rapidly advanced the field. Yet, heterogeneous training setups across studies make it difficult to determine whether performance improvements came from architectural innovations or effective data engineering. In this work, we investigate whether a standard patch Transformer architecture, paired with a straightforward training protocol, can match the zero-shot forecasting performance of state-of-the-art TSFMs. We conduct a comprehensive ablation study covering model scaling, data composition, and training techniques to examine the ingredients behind this performance. Our findings show that pretraining data composition, contiguous patch masking, and context length have a substantial effect on forecasting accuracy, and that the generic architecture itself can deliver high performance with good scalability. By strictly controlling these experimental variables, we provide comprehensive empirical results on model scaling across multiple dimensions. Furthermore, we establish that the effectiveness of this training protocol generalizes robustly across benchmarks and alternative architectures. We release our pretrained models and document the detailed training protocol and findings to establish a transparent, reproducible baseline for future research.
\end{abstract}

\section{Introduction}


The recent shift from domain-specific models to cross-domain Time Series Foundation Models (TSFMs) capable of zero-shot generalization marks a significant advancement in the field. These models have demonstrated remarkable success across a diverse range of domains and downstream tasks \citep{godahewa2021monash, jiang2023libcity, emami2023buildingsbench, liu2023largest, nguyen2023climatelearn}, achieving performance that matches or exceeds state-of-the-art (SOTA) supervised methods \citep{goswami2024moment, wen2024abstracted, shi2025timemoe}. Given the ubiquity of time series data and the proven feasibility of this approach, TSFMs have garnered substantial attention in the recent literature.

The performance of a foundation model is fundamentally the joint product of its backbone architecture and its training protocol. However, as the field of TSFMs rapidly expands, a concerning trend has emerged: many recent studies claim advantage in architecture while evaluating models trained using highly heterogeneous datasets and processing techniques. This lack of standardization introduces three critical challenges for the community. First, it creates a confounding factor, making it difficult to attribute performance improvements to actual architectural innovations rather than superior data engineering. Second, although many models are open-sourced, their training protocols are often proprietary or under-documented, severely limiting reproducibility. Finally, for researchers developing new models, the absence of a proven, transparent training recipe makes it nearly impossible to determine whether a novel architectural idea is inherently flawed or simply trained incorrectly. Consequently, the true impact of recent architectural advancements remains obscured.


\begin{figure}[t]
    \centering
    \begin{minipage}[t]{0.56\textwidth}
        \vspace{0pt}
        \centering
        \includegraphics[width=\linewidth]{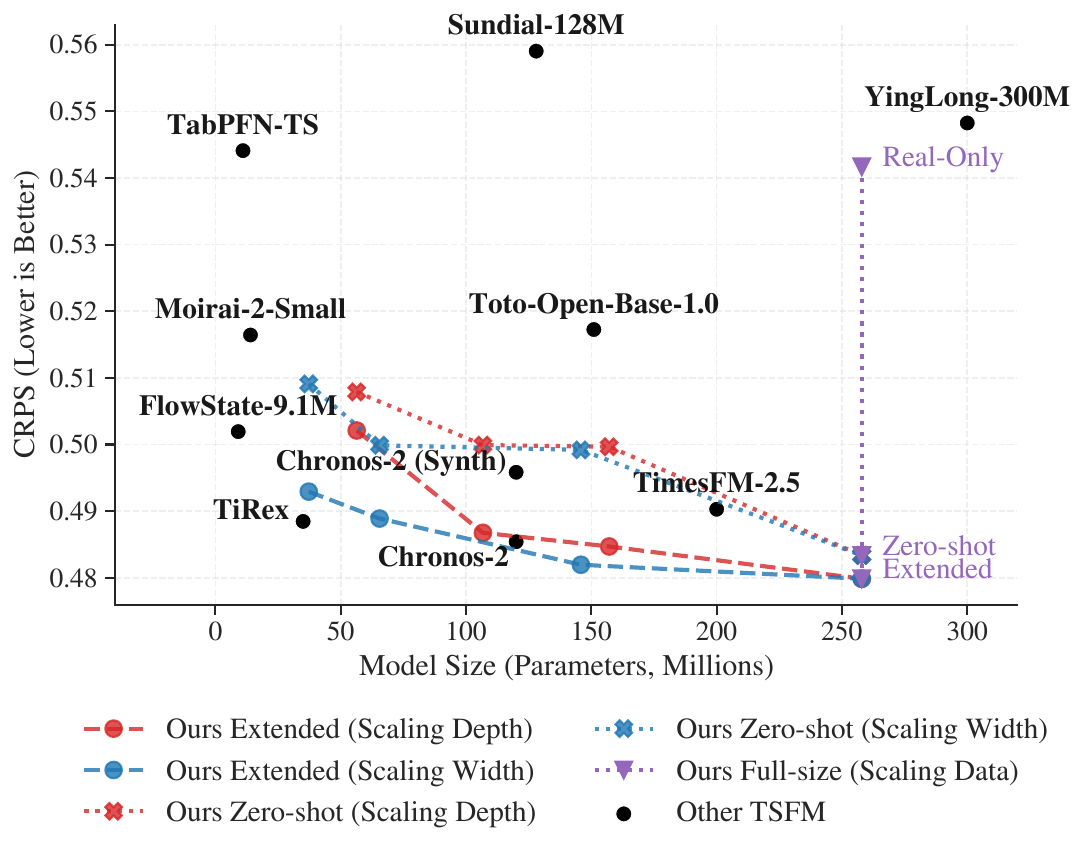}
        \caption{\textbf{The Landscape of Time Series Foundation Models on Probabilistic Forecasting}. We compare our generic Transformer with SOTA TSFMs on the GIFT-Eval benchmark. The results establish a strong reference for model scaling in TSFMs and show that, under our protocol, enriching the pretraining data (purple line) yields larger gains than scaling the model size. The Extended model is trained on a corpus that overlaps 23 GIFT-Eval test cases (Section \ref{sec:exp-setup}).}
        \label{fig:size-crps}
    \end{minipage}\hfill
    \begin{minipage}[t]{0.42\textwidth}
        \vspace{0pt}
        \centering
        \includegraphics[width=\linewidth]{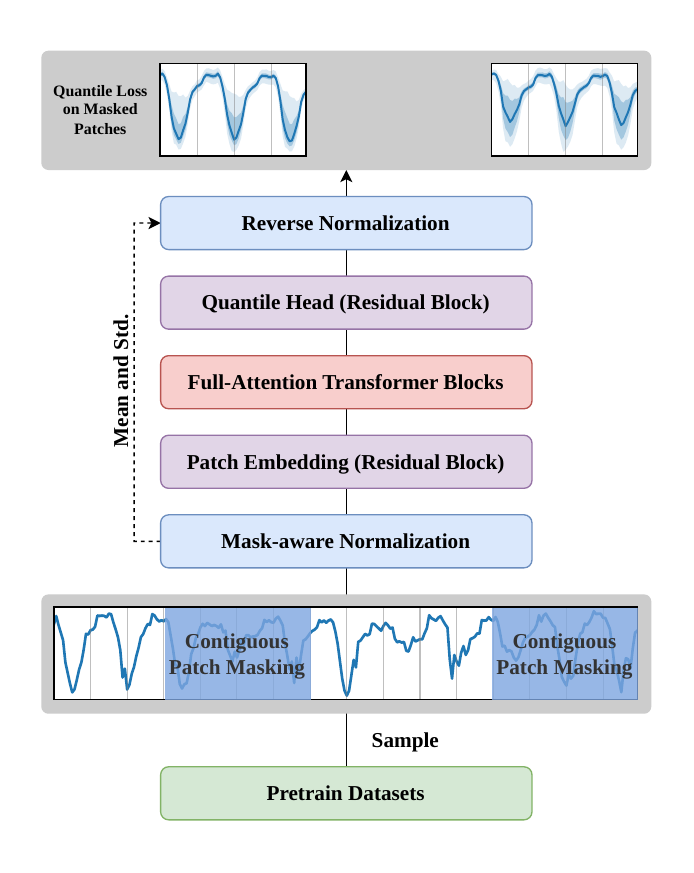}
        \vspace{-0.9cm}
        \caption{\textbf{An overview of the generic Transformer architecture}, an encoder-only Transformer with patched inputs and quantile outputs.}
        \label{fig:architecture}
    \end{minipage}
\end{figure}

In this work, we demonstrate that a generic patch Transformer architecture, when coupled with a straightforward training recipe, achieves zero-shot performance on par with the leading TSFMs on the GIFT-Eval probabilistic forecasting benchmark \citep{aksu2025gifteval} and the TIME benchmark \citep{qiao2026time}. Following this finding, we investigate the fundamental question: \emph{What are the essential ingredients of this effectiveness?} To answer this, we conduct a comprehensive empirical study to examine the contribution of each ingredient required for a successful TSFM.

As illustrated in Figure \ref{fig:size-crps}, we compare the scaling behavior of our generic patch Transformer against recent SOTA TSFMs. We observe that this standard architecture delivers performance comparable to, and sometimes exceeding, deliberately designed TSFMs across various model sizes. Notably, within this controlled setting, enriching the pretraining data composition yields larger gains than scaling the model size, highlighting the central role of data in TSFM performance.

\textbf{Our contributions} are summarized as follows:
\begin{itemize}[nosep, leftmargin=0.3cm]
    \item \textbf{Competitive Performance with a Generic Architecture}: We demonstrate that a standard Transformer architecture, utilizing patched inputs and quantile outputs, achieves zero-shot performance on the GIFT-Eval and TIME benchmarks that is statistically indistinguishable from that of the leading TSFMs (Section \ref{sec:benchmark}, Appendix \ref{append:stats}). This result serves as strong evidence for the effectiveness of generic, scalable backbones.
    \item \textbf{Comprehensive Component Analysis}: We conduct a systematic ablation study to examine the impact of critical training components, including model scaling laws, pretraining data composition, and training techniques. With comprehensive empirical experiments on each aspect, we unveil the essential ingredients for the success of TSFM.
    \item \textbf{Key Insights and Analysis}: Our analysis identifies pretraining data composition as an essential factor where both real-world and synthetic datasets contribute to a successful TSFM. We also provide extensive empirical evidence detailing how other key aspects affect the zero-shot performance of TSFM, including model scaling, contiguous patch masking, context length.
    \item \textbf{Open Research}: We release our pretrained model checkpoints and model implementation\footnote{Model checkpoints: \url{https://huggingface.co/ibm-research/patchtst-fm-r1}; model code: \url{https://github.com/ibm-granite/granite-tsfm/tree/patchtst-fm}.}. We further document the complete training protocol, including pretraining data composition and sampling, masking strategy, optimization settings, and compute (Appendix \ref{append:framework}), together with detailed empirical results and statistical analyses. By providing a transparent and reproducible baseline built from publicly available data sources, we aim to facilitate fair comparisons and accelerate future research in TSFMs.
\end{itemize}

\textbf{Scope of this Work}:
To ensure clarity and rigor, we define the scope of our study as follows:
\begin{itemize}[nosep, leftmargin=0.3cm]
    \item \textbf{Univariate Modeling and Zero-shot Forecasting:} Our primary evaluation metric is zero-shot performance on univariate time series forecasting tasks. Specifically, our study focuses on the recipe of training a high-quality base foundation model. Therefore, no adaptation approaches, such as finetuning or agentic methods, are compared.
    \item \textbf{Ablation of Intrinsic Factors:} We focus exclusively on analyzing the scaling behaviors and component contributions within our proposed generic architecture. Rather than attempting to reproduce and retrain alternative architectures, which introduces significant variance due to differing implementations and resource requirements, we prioritize a controlled, in-depth analysis of the standard Transformer to establish a robust baseline for the field.
\end{itemize}

\section{Architecture} \label{sec:architecture}

In this section, we present the formulation of our Time Series Foundation Model (TSFM) architecture, as illustrated in Figure \ref{fig:architecture}. Our design leverages a standard Transformer backbone augmented with specific data processing and training techniques designed to enhance zero-shot generalization.

\subsection{Problem Formulation}

Let $\mathbf{x} \in \mathbb{R}^T$ denote a univariate input time series of length $T$. To perform probabilistic predictions, our architecture takes $\mathbf{x}$ as input and predicts a set of quantiles $\hat{\mathbf{q}} \in \mathbb{R}^{T \times K}$ representing the distribution of the target values. This enables the model to capture the inherent uncertainty in time series data via a deterministic mapping.

To form semantically-meaningful tokens from real-value inputs, the time series $\mathbf{x}$ is segmented into non-overlapping patches $\mathbf{p} \in \mathbb{R}^{N \times L}$, where $L$ is the patch size and $N = \frac{T}{L}$ denotes the number of patches \citep{nie2023patchtst}. Note that we pad or truncate the raw inputs to make a constant length $T$ that is always divisible by $L$.

\subsection{Masking for Long-Horizon Prediction Training}


When modeling real-valued continuous systems with neural networks, conventional training paradigms—such as single, step next-token prediction or scattered masked autoencoding, often fail to capture long-horizon dynamics. To mitigate this, various disciplines have adopted strategies that explicitly optimize for multi-step predictions over extended horizons \citep{hafner2019rssm, gloeckle2024mtpllm}. Within the context of TSFMs, an instantiation of this principle is termed Contiguous Patch Masking (CPM) \citep{auer2025tirex}.

We define a binary mask $\mathbf{m} \in \{0, 1\}^T$, where $m_t=1$ indicates a masked observation and $m_t=0$ indicates a visible observation. Unlike the conventional approaches that uniformly sample scattered masks \citep{he2022mae, woo2024moirai}, CPM selects random starting positions (multiples of $L$) and masks blocks of $N_{\text{cpm}} \cdot L$ consecutive observations, as illustrated by the blue shaded regions in Figure \ref{fig:architecture}. This creates substantial ``information gaps'' between the visible context and the targets, compelling the model to learn the underlying dynamics of the time series to bridge these gaps.





\subsection{Patch Embedding and Transformer Backbone}

We first apply a mask-aware variant of Reversible Instance Normalization \citep{kim2022revin} that computes statistics only over visible time points to prevent leakage from masked targets (see Appendix \ref{sec:normalization} for details). The normalized patches $\bar{\mathbf{p}}$ and the masking $\mathbf{m}$ are projected into latent embeddings $\mathbf{h} \in \mathbb{R}^{N \times d}$ using a residual block \citep{das2024timesfm, auer2025tirex, ansari2025chronos2}, defined as 
\begin{equation}
    \mathbf{h}_{\text{out}} = f_{\text{res}}(\mathbf{h}_{\text{in}}) + f_{\text{out}}\big( \texttt{sigmoid(}f_{\text{in}}(\mathbf{h_{\text{in}}})) \big),
\end{equation}
where $f_{\text{res}}$, $f_{\text{in}}$, $f_{\text{out}}$ are linear layers. Learnable positional embeddings are then added to these patch embeddings. The resulting sequence of embeddings is processed by a stack of full-attention Transformer blocks (i.e., an encoder-only Transformer) with depth $N_{\text{layer}}$ and embedding size $d$. We adhere to the standard Transformer formulation with pre-normalization.

\subsection{Probabilistic Prediction}

To generate probabilistic predictions, we employ a Quantile Head \citep{auer2025tirex, ansari2025chronos2}, implemented as a residual block, which maps the latent representations to the quantile distribution of the patches. Each patch embedding $\mathbf{h}_i \in \mathbb{R}^{d}$ is decoded into the normalized quantile values of the corresponding patch $\hat{\mathbf{q}}_i \in \mathbb{R}^{L \times K}$. The predictions for all patches are concatenated into $\hat{\mathbf{q}} \in \mathbb{R}^{T \times K}$. For each $q_t \in \mathbb{R}^K$, the model predicts the values for a set of quantile levels $Q = \{ \tau_1, \tau_2, \dots, \tau_K \}$.

During inference, the predicted quantiles are scaled back using the stored statistics with $\mathbf{y} = \mathtt{sinh}(\hat{\mathbf{q}}) \cdot \sigma + \mu$.

\subsection{Training Objective}

The model is trained to minimize the Pinball Loss (Quantile Loss), computed exclusively on the masked real values. Formally, the objective function is defined as:
\begin{equation} \label{eq:loss}
\begin{split}
    & \mathcal{L}(\hat{\mathbf{q}}, \mathbf{x}) = \frac{1}{\sum_{t=1}^T \mathds{1}(m_t = 1)} \sum_{t=1}^{T} \mathds{1}(m_t = 1) \mathcal{L}_q(\hat{\mathbf{q}}_t, x_t), \\
    & \text{where} \hspace{0.2cm} \mathcal{L}_q(\hat{\mathbf{q}}_t, x_t) = \sum_{\tau \in Q} (x_t - \hat{q}_t^{(\tau)}) \cdot (\tau - \mathds{1}(x_t \leq \hat{q}_t^{(\tau)})).
\end{split}
\end{equation}
Here, $\hat{q}_t^{(\tau)}$ denotes the predicted value for quantile level $\tau$ at timestamp $t$.

\subsection{Design Analysis}
Our architectural design is guided by the following core principles:
\begin{itemize}[nosep, leftmargin=0.3cm]
    \item \textbf{Simplicity}: We deliberately adhere to the generic Transformer formulation, demonstrate its effectiveness for TSFMs, and study the scaling laws in Section \ref{sec:ablation}.
    \item \textbf{Flexibility}: Our use of CPM with randomized positions and high masking ratios prevents the model from overfitting to a specific prediction horizon, which results in flexible forecast lengths (see Figure \ref{fig:synth_pred_len} and Appendix \ref{append:horizon}).
    \item \textbf{Single-step Inference}: The encoder-only architecture enables parallel decoding of all target patches simultaneously. Unlike autoregressive models that generate predictions sequentially, our approach produces the entire forecast horizon in a single forward pass. This minimizes the propagation of prediction error or model's inherent bias, and thus maximizes the role of real context from data. (See Appendix \ref{append:speed}.)
\end{itemize}

\section{Empirical Evaluations}

\begin{figure*}
    \centering
    \includegraphics[width=\linewidth]{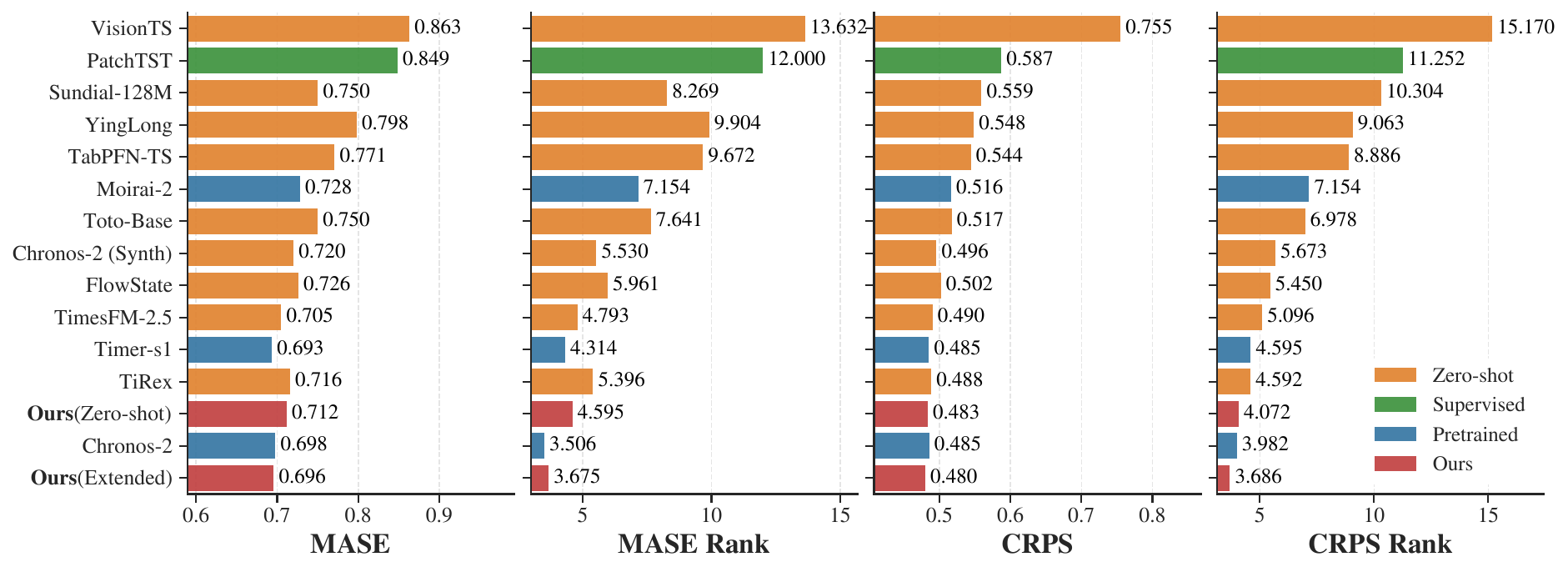}
    \caption{Forecasting performance on the GIFT-Eval benchmark. The four metrics (MASE, MASE Rank, CRPS, CRPS Rank) are aggregated over the 97 test cases. Lower values indicate better performance. The methods are sorted based on CRPS Rank. Ours (Extended) is trained on a corpus that overlaps 23 of the 97 test cases and is therefore not a zero-shot result (Section \ref{sec:exp-setup}).}
    \label{fig:gift-benchmark}
\end{figure*}

In this section, we rigorously evaluate the performance of our proposed architecture. We begin by detailing the experimental setup, including model hyperparameters and the composition of our pretraining corpus. Then, we demonstrate the performance of our models on probabilistic forecasting on a public benchmark.

\subsection{Experiment Setups} \label{sec:exp-setup}

\textbf{Model Configuration}: The default hyperparameters, presented in Table \ref{tab:hyperparameters}, are selected heuristically based on established best practices in the Transformer literature. The detailed implementations are summarized in Appendix \ref{append:model-details}, and Appendix \ref{append:design-summary} summarizes the empirical evidence behind each choice.

\textbf{Pretrain Data}: A critical component of our study is the curation of a diverse and high-volume pretraining corpus. Our dataset comprises three sources:
\begin{itemize}[nosep, leftmargin=0.3cm]
    \item \textbf{Real Data}: We utilize the GIFT-Eval-Pretrain collection, consisting of 132 datasets from various domains totaling approximately 4.5 million time series and 230 billion observations.
    \item \textbf{KernelSynth} (Synthetic): To support long-context training ($T=8192$), we generate 10 million univariate time series using the KernelSynth procedure \citep{ansari2024chronos}. As suggested by \citet{auer2025tirex}, we employ a diverse set of periodic kernels and a small volume of augmentations.
    \item \textbf{TSMixup} (Augmented): To enhance temporal diversity, we incorporate TSMixup data \citep{ansari2024chronos}, generated by randomly mixing real-world time series. We evaluate two variants to study the impact of data leakage: (1) a ``Leaky'' set of 10 million series that includes series drawn from datasets represented in 23 of the 97 GIFT-Eval test cases, and (2) a ``Clean'' set of 4 million series with all test-set overlaps removed. This comparison allows us to quantify the true zero-shot capability of the model versus performance gains from domain exposure.
\end{itemize}

\textbf{Evaluated Models}: Based on these data collections, we present results for two distinct model variants:
\begin{itemize}[nosep, leftmargin=0.3cm]
    \item \textbf{Zero-shot Model}: Trained on Real Data + 10M KernelSynth + 4M Clean TSMixup. This model has no exposure to any test distribution and is used for all zero-shot comparisons on GIFT-Eval.
    \item \textbf{Extended Model}: Trained on Real Data + 10M KernelSynth + 10M Leaky TSMixup. Because its corpus overlaps 23 GIFT-Eval test cases, its GIFT-Eval results are not zero-shot; we report them as a diagnostic of the effect of broader data coverage and domain exposure, mark them with $^\dagger$ in the tables, and note the overlap in the figure captions. On the TIME benchmark, none of the 98 cases is seen by either variant, so both models are evaluated zero-shot there.
\end{itemize}

\subsection{Benchmark Results} \label{sec:benchmark}

To assess zero-shot probabilistic forecasting performance, we utilize the GIFT-Eval benchmark \citep{aksu2025gifteval}, a widely adopted standard for TSFMs. The benchmark encompasses 55 diverse real-world datasets spanning 97 distinct test cases, covering multiple domains of time series data and various forecast lengths.

\textbf{Compared Methods}: We compare our approach against the current SOTA TSFMs on the GIFT-Eval leaderboard, selecting the most recent high-performing TSFM variants, including TabPFN-TS \citep{hoo2024tabpfnts}, VisionTS \citep{chen2025visionts}, Sundial \citep{liu2025sundial}, Moirai-2 \citep{woo2024moirai}, Toto \citep{cohen2025toto}, YingLong \citep{wang2025yinglong}, FlowState \citep{graf2025flowstate}, TimesFM-2.5 \citep{das2024timesfm}, TiRex \citep{auer2025tirex}, Timer-S1 \citep{liu2026timers1}, and Chronos-2 \citep{ansari2025chronos2}. To provide a comprehensive context, we also include two statistical baselines (Naive and Seasonal Naive) and a supervised deep learning baseline PatchTST \citep{nie2023patchtst}. The baseline results are obtained from the official GIFT-Eval Leaderboard. Detailed information about the baselines is summarized in Appendix \ref{append:baselines}.

\textbf{Note on Comparability}: It is important to emphasize that the baseline TSFMs utilize heterogeneous training setups, varying significantly in pretraining data scale, composition, model configurations, and optimization recipes. Consequently, the benchmarking results below reflect the aggregate performance of the final model artifacts (the combination of architecture, data, and training) rather than the isolated efficacy of the architectures themselves. Our goal is not to claim architectural superiority, but to demonstrate that a generic architecture is sufficient to match the leading TSFMs when coupled with a robust training recipe.

\textbf{Evaluation Metrics}: Following the standard GIFT-Eval protocol, we report three key metrics:
\begin{itemize}[nosep, leftmargin=0.3cm]
    \item \textbf{Mean Absolute Scaled Error (MASE)}: Evaluates point forecasting accuracy by scaling the Mean Absolute Error of the predicted median (i.e., the $50^{\text{th}}$ quantile $\hat{\mathbf{q}}^{(0.5)}$) against the error of a naive forecast.
    \item \textbf{Continuous Ranked Probability Score (CRPS)}: Assesses probabilistic forecasting quality by measuring the difference between the predicted cumulative distribution function and the ground truth observation. It effectively generalizes MAE to the probabilistic setting.
    \item \textbf{Rank}: Measures the relative standing of each method among all the compared methods above. Rankings are computed per-dataset based on MASE and CRPS and averaged over all cases. The models are sorted by CRPS Rank as our primary focus on probabilistic forecasting.
\end{itemize}
Then, on each test dataset and case, the average MASE and CRPS over all test samples are normalized w.r.t. the performance of Seasonal Naive. The metrics over the 97 cases are aggregated with geometric mean for MASE and CRPS, and arithmetic mean for Rank, as an overall evaluation of the methods on the GIFT-Eval benchmark (refer to Appendix \ref{append:gift-eval} and \citet{aksu2025gifteval} for more details).

\textbf{Results Analysis}: Figure \ref{fig:gift-benchmark} summarizes the comparative performance on the GIFT-Eval benchmark. Our Zero-shot model ranks among the top TSFMs on all metrics. A paired bootstrap analysis over test cases (Appendix \ref{append:stats}) shows that its aggregate MASE and CRPS are statistically indistinguishable from those of the strongest baselines, Chronos-2, TimesFM-2.5, and Timer-S1: all 95\% confidence intervals of the score ratios include 1. The Extended model obtains the best MASE and CRPS point estimates; since its corpus overlaps 23 test cases, we interpret this result as a diagnostic of domain exposure rather than as a zero-shot result (Section \ref{sec:ablation-data}). These findings provide strong empirical evidence that a generic Transformer architecture, when paired with a robust training recipe, is sufficient to match the leading TSFMs on time-series forecasting. Furthermore, our results demonstrate that the effective utilization of existing open-source data can yield models that match the recent TSFMs which rely on proprietary synthetic generation pipelines.

\section{Systematic Ablation Study} \label{sec:ablation}

Recognizing that the generic Transformer architecture can match the leading TSFMs on the GIFT-Eval benchmark, we now turn to the critical question: \textit{What are the essential ingredients for this success?} As discussed in the previous sections, the heterogeneous, sometimes even opaque, training recipes of existing TSFMs often obscure whether improvements stem from architectural innovations or simply from better data and training protocols.

In this section, we conduct a large-scale ablation study to systematically break down these factors. We vary one component at a time and analyze its contribution. By strictly controlling variables within our standardized training pipeline, we aim to provide a transparent, empirical ``recipe'' for training effective TSFMs.

Unless stated otherwise, the ablations in this section start from the Extended configuration and are evaluated on all 97 GIFT-Eval cases; they compare training choices within a single recipe and are not zero-shot results (Section \ref{sec:exp-setup}). Throughout this section, the Rank metric is the CRPS Rank computed strictly within the context of each specific ablation experiment. For instance, when comparing a set of ablation variants, their ranks are derived by comparing them against each other and the fixed set of baseline methods from Section \ref{sec:benchmark}. Consequently, the reported Rank values are relative measures intended for comparison within a specific table or figure, and should not be compared directly across different ablation studies.

\subsection{Pretrain Data Composition} \label{sec:ablation-data}

Table \ref{tab:dataset_ablation} and the purple line in Figure \ref{fig:size-crps} 
investigate the impact of training data diversity on downstream performance. All variants share the same training steps, context length, and effective batch size, i.e., the same nominal token budget. We conclude the following:
\begin{enumerate}[nosep, leftmargin=0.3cm]
    \item \textbf{Insufficiency of Single Source}: Training on KernelSynth alone yields the poorest results, indicating that while synthetic periodic kernels capture basic seasonality, they fail to model complex real-world noise and non-stationary dynamics. Using only real data provides a stronger baseline, yet performance remains suboptimal.
    \item \textbf{The ``Synergy'' Effect}: The Zero-shot configuration, combining real and synthetic data, delivers a substantial performance boost. This suggests a complementary relationship: synthetic data acts as a robust regularizer, filling the gaps between diverse real-world distributions, while the real data provides patterns that cannot be easily synthesized.
    \item \textbf{Benefits of Domain Exposure}: The Extended model, which includes the broader leaky TSMixup set, achieves the best point estimates on GIFT-Eval. While the architecture generalizes well to unseen distributions (the Zero-shot variant), exposing the model to the target domain's patterns, even via mixed augmentations, provides an additional performance edge.
\end{enumerate}

\textbf{Quantifying the Impact of Leakage}: The performance gap between our variants raises a critical question: \textit{does the leaked data merely allow the model to memorize specific test cases, or does it contribute to better feature learning?} To investigate this, we excluded all 23 leaked test cases from the benchmark and compared the models exclusively on the remaining 74 unseen cases. As illustrated in Figure \ref{fig:gift-benchmark-no-leak}, both of our variants remain among the top models on this non-overlapping subset. Notably, the gap between the Zero-shot and Extended variants narrows, and even reverses slightly on the overall CRPS. To determine whether the increased data diversity still yields genuine zero-shot improvements, we evaluated both models on the TIME benchmark \citep{qiao2026time}, where no test case is seen by either model. As detailed in Table \ref{tab:time-benchmark}, the Extended model outperforms the Zero-shot model on both metrics, and the 95\% intervals of the Extended/Zero-shot score ratios exclude 1 (Appendix \ref{append:stats}). Together, these results support our core finding: pretraining data composition has a substantial effect on benchmark performance, and expanding the diversity of the pretraining corpus improves generalization to unseen datasets, although these comparisons vary volume, composition, and domain exposure jointly and do not isolate any single factor.


\begin{figure}[t]
    \centering
    \begin{minipage}{0.49\textwidth}
        \centering
        \includegraphics[width=\linewidth]{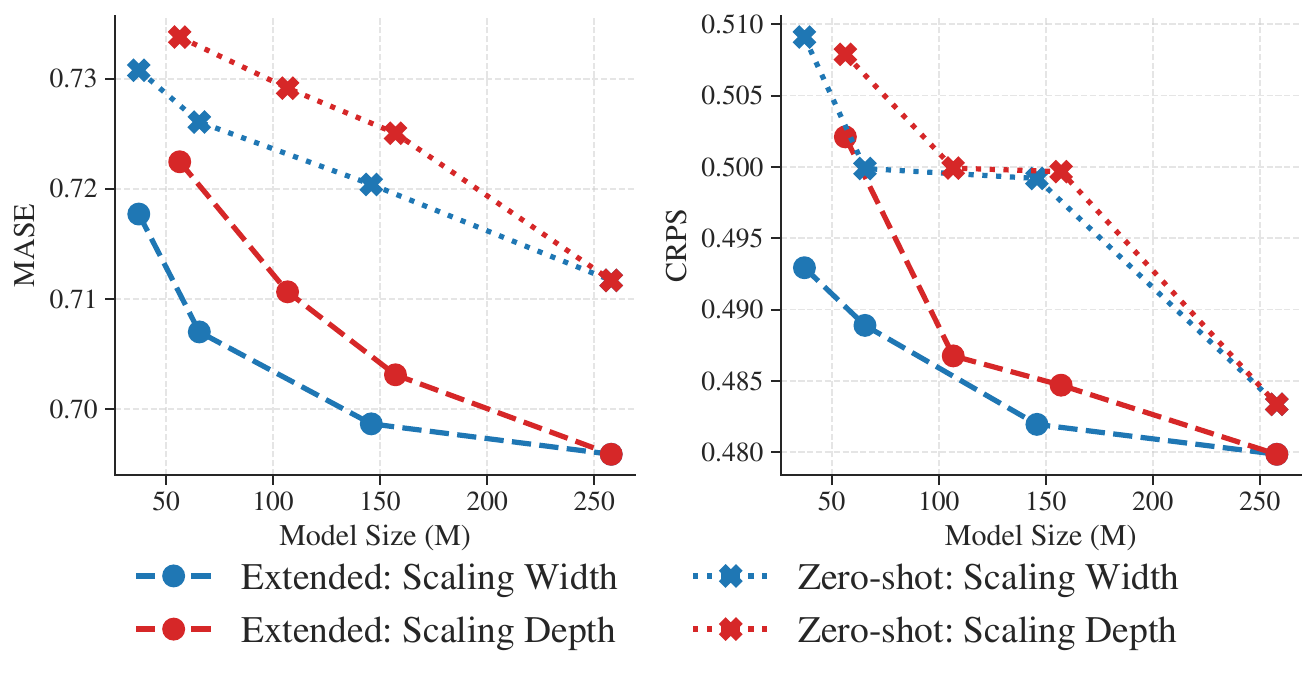}
        \caption{\textbf{Impact of model scaling on forecasting performance}: We compare increasing model capacity via embedding dimension (width) versus layer count (depth) for both the Extended and Zero-shot corpora (settings in Appendix \ref{append:scaling-details}).}
        \label{fig:model-scaling}
    \end{minipage}\hfill
    \begin{minipage}{0.49\textwidth}
        \centering
        \captionof{table}{Ablation study on \textbf{context length}. Extending the context yields significant improvements across all metrics. 
        Models pretrained with longer context lengths provide more robust and general usage across diverse forecast lengths.}
        \label{tab:context_ablation}
        \vspace{0.3cm}
        \begin{tabular}{lccc}
        \toprule
        \textbf{Ctx. Len.} & \textbf{MASE} $\downarrow$ & \textbf{CRPS} $\downarrow$ & \textbf{Rank} $\downarrow$ \\
        \midrule
        2048 & 0.721 & 0.500 & 4.592 \\
        4096 & 0.717 & 0.497 & 4.501 \\
        \rowcolor{gray!15} \textbf{8192} & \textbf{0.696} & \textbf{0.480} & \textbf{3.381} \\
        \bottomrule
        \end{tabular}
    \end{minipage}
\end{figure}

\subsection{Transformer Backbone Dimensions}

We investigate the impact of model capacity on forecasting accuracy by independently scaling the Transformer backbone along two axes: (1) keeping $N_\text{layer}=20$ and scaling the embedding dimension $d \in \{384, 512, 768, 1024\}$, and (2) keeping $d=1024$ and scaling the model depth $N_\text{layer} \in \{4, 8, 12, 20\}$. Figure \ref{fig:model-scaling} illustrates the scaling behavior across both MASE and CRPS metrics. We observe two key trends:
\begin{enumerate}[nosep, leftmargin=0.3cm]
    \item \textbf{Predictable Monotonic Scaling}: Performance improves consistently as parameter count increases, in line with the scaling behavior observed in other domains. This also suggests that the current architecture has not yet saturated and could benefit from further scaling.
    \item \textbf{Depth versus Width}: At similar parameter counts, deep-narrow models consistently outperform shallow-wide models. This indicates that univariate TSFM benefits more from the hierarchical reasoning capabilities of a deep Transformer stack than from the increased feature dimension. Consequently, for parameter-efficient designs, prioritizing network depth yields better performance returns than expanding embedding size.
\end{enumerate}
These sweeps characterize the accuracy--size trade-off within the Transformer backbone (see Appendix \ref{append:scaling-details} for the exact settings) rather than arguing for larger models per se, since larger models also incur higher training and inference cost.

\subsection{Context Length}

We evaluate the importance of historical information by varying the input context length $T \in \{2048, 4096, 8192\}$. Table \ref{tab:context_ablation} demonstrates a clear benefit of scaling the lookback window on the general forecasting performance. However, we observe that these gains are most pronounced in time series that exhibit long-term seasonality (i.e., where the seasonal period is large relative to the sampling rate). In such cases, shorter lookback windows often lead to ambiguity between periodic patterns and trends, leading to suboptimal predictions. (See Figure \ref{fig:context-compare} for examples)

\subsection{Pretrain Masking Strategy}

\begin{table}[h]
    \centering
    
    \begin{minipage}[t]{0.48\textwidth}
        \centering
        \caption{Impact of \textbf{Contiguous Patch Masking (CPM) block size}. The results show a significant performance improvement from CPM over scattered masking (i.e., $N_{\text{CPM}}=1$).}
        \label{tab:cpm_ablation}
        \begin{tabular}{lccc}
            \toprule
            \textbf{Num. CPM} & \textbf{MASE} $\downarrow$ & \textbf{CRPS} $\downarrow$ & \textbf{Rank} $\downarrow$ \\
            \midrule
            1 & 0.750 & 0.514 & 6.675 \\
            4 & 0.704 & 0.482 & 3.790 \\
            \rowcolor{gray!15} \textbf{8} & \textbf{0.696} & \textbf{0.480} & \textbf{3.566} \\
            \bottomrule
        \end{tabular}
    \end{minipage}\hfill
    \begin{minipage}[t]{0.48\textwidth}
        \centering
        \caption{Impact of \textbf{Masking Ratio}. The architecture remains relatively robust to this hyperparameter.}
        \label{tab:mask_ratio_ablation}
        \begin{tabular}{lccc}
            \toprule
            \textbf{Mask Ratio} & \textbf{MASE} $\downarrow$ & \textbf{CRPS} $\downarrow$ & \textbf{Rank} $\downarrow$ \\
            \midrule
            20\% & 0.698 & 0.484 & \textbf{3.900} \\
            \rowcolor{gray!15} \textbf{40\%} & \textbf{0.696} & \textbf{0.480} & 3.972 \\
            60\% & 0.702 & 0.484 & 4.420 \\
            80\% & 0.703 & 0.482 & 4.678 \\
            \bottomrule
        \end{tabular}
    \end{minipage}

\end{table}

We analyze the design of the training objective by ablating the masking strategy, which serves as the primary mechanism for masked autoencoding, self-supervised learning. Based on results in Table \ref{tab:cpm_ablation} and Table \ref{tab:mask_ratio_ablation}, we conclude that:
\begin{itemize}[nosep, leftmargin=0.3cm]
    \item \textbf{Necessity of Training for Long-term Predictions}: Table \ref{tab:cpm_ablation} investigates the impact of the CPM block length $N_{\text{CPM}} \in \{1, 4, 8\}$. We observe that standard random patch masking (i.e., $N_{\text{CPM}}=1$) yields poor performance. In contrast, introducing CPM significantly improves the performance by creating large ``information gaps'' and forcing the model to learn the underlying system dynamics for long-term predictions. Additionally, using larger $N_{\text{CPM}}$ yields a performance edge by further pushing the model towards long-horizon understanding. (See Figure \ref{fig:cpm-compare} for examples.)
    \item \textbf{Robustness to Mask Ratio}: Table \ref{tab:mask_ratio_ablation} explores the optimal density of the masking signal. We find that the architecture is generally robust to the masking ratio, with no catastrophic failure observed even at high ratios (i.e., $80\%$). A ratio of $40\%$ emerges as a ``sweet spot'', achieving slightly better MASE and CRPS.  
\end{itemize}

\subsection{Generalization to Other Benchmarks and Architectures} \label{sec:others}

\begin{table}[ht]
\centering
\begin{minipage}[t]{0.48\linewidth}
    \centering
    \caption{\textbf{Generalization to novel benchmarks.} Forecasting performance of our generic Transformer on the TIME benchmark, compared against SOTA TSFMs. Metrics are aggregated across 50 distinct datasets and 98 test cases.}
    \label{tab:time-benchmark}
    \resizebox{\linewidth}{!}{%
    \begin{tabular}{lcccc}
    \toprule
    & \multicolumn{2}{c}{\textbf{MASE} $\downarrow$} & \multicolumn{2}{c}{\textbf{CRPS} $\downarrow$} \\
    \cmidrule(lr){2-3} \cmidrule(lr){4-5}
    \textbf{Model} & \textbf{Score} & \textbf{Rank} & \textbf{Score} & \textbf{Rank} \\
    \midrule
    Chronos-2                 & \textbf{0.660} & 3.929          & \textbf{0.549} & \textbf{3.796} \\
    \rowcolor{gray!15} \textbf{Ours (Extended)}  & \textbf{0.660} & \textbf{3.776} & \textbf{0.549} & 4.071 \\
    \rowcolor{gray!15} \textbf{Ours (Zero-shot)} & 0.669          & 5.224          & 0.556          & 4.724 \\
    TimesFM-2.5               & 0.667          & 5.112          & 0.559          & 5.510 \\
    Timer-S1                  & 0.674          & 5.612          & 0.565          & 5.694 \\
    TiRex                     & 0.681          & 6.286          & 0.565          & 5.827 \\
    Toto                      & 0.696          & 8.408          & 0.576          & 7.643 \\
    Moirai2 & 0.701 & 8.735 & 0.580 & 8.204 \\
    \bottomrule
    \end{tabular}%
    }
\end{minipage}\hfill 
\begin{minipage}[t]{0.5\linewidth}
    \centering
    \caption{\textbf{Generalization across architectures.} Forecasting performance on the GIFT-Eval benchmark comparing top-tier TSFMs against four generic architectures trained with our training protocol (shaded rows).}
    \label{tab:gift-architecture}
    \resizebox{\linewidth}{!}{%
    \begin{tabular}{llcccc}
    \toprule
    & & \multicolumn{2}{c}{\textbf{MASE} $\downarrow$} & \multicolumn{2}{c}{\textbf{CRPS} $\downarrow$} \\
    \cmidrule(lr){3-4} \cmidrule(lr){5-6}
    \textbf{Model} & \textbf{Size} & \textbf{Value} & \textbf{Rank} & \textbf{Value} & \textbf{Rank} \\
    \midrule
    \rowcolor{gray!15} Transformer       & 33M  & 0.731 & 7.766  & 0.509 & 7.368 \\
    Toto              & 151M  & 0.750 & 8.181  & 0.517 & 7.544 \\
    Moirai2           & 14M   & 0.728 & 7.576  & 0.516 & 7.906 \\
    \rowcolor{gray!15} MLP-Mixer         & 30M  & 0.745 & 9.259  & 0.518 & 8.718 \\
    \rowcolor{gray!15} Bi-Mamba2         & 37M  & 0.755 & 9.517  & 0.518 & 8.994 \\
    TabPFN-TS         & 11M   & 0.771 & 10.856 & 0.544 & 9.911 \\
    YingLong          & 300M & 0.798 & 11.113 & 0.548 & 10.144 \\
    Sundial           & 128M   & 0.750 & 9.103  & 0.559 & 11.711 \\
    \rowcolor{gray!15} CNN-ResNet        & 30M  & 0.791 & 12.447 & 0.555 & 12.297 \\
    \bottomrule
    \end{tabular}%
    }
\end{minipage}
\end{table}

We demonstrate that our results are not limited to the GIFT-Eval benchmark. We evaluate our generic Transformer on the TIME benchmark, which includes 50 fresh datasets for zero-shot evaluation. As shown in Table \ref{tab:time-benchmark}, our models remain comparable with the leading TSFMs (bootstrap intervals in Appendix \ref{append:stats}). This indicates that our training recipe yields robust models that transfer well to unseen domains and datasets.

Finally, we test if the training protocol generalizes to other backbone architectures. With the zero-shot protocol, we train three additional generic models that represent different types of basic inductive biases: MLP-Mixer \citep{ekambaram2023mlpmixer}, bidirectional Mamba2 (Bi-Mamba2) \citep{dao2024mamba}, and a CNN-ResNet. All four backbones share the data recipe, training steps, context length, and batch size (i.e., the same nominal token budget) and are roughly matched in parameter count but not in FLOPs; each is a single run (Appendix \ref{append:scaling-details}). Table \ref{tab:gift-architecture} details their performance on the GIFT-Eval benchmark. Notably, when coupled with our training protocol, even the very basic architectures could match or surpass deliberately designed TSFMs. This supports our core claim: a rigorous training protocol can matter as much as architectural innovation for a foundation model's empirical success, and performance differences between models trained under different data and protocols cannot be attributed to architecture alone.

\section{Discussion and Conclusion}

Figure \ref{fig:size-crps} visualizes the broader landscape of the capability of existing TSFMs on probabilistic forecasting, synthesizing our benchmarking results in Section \ref{sec:benchmark} with the scaling dynamics observed in our ablations in Section \ref{sec:ablation}. By comparing the performance trajectories of our generic Patch Transformer architecture against the current SOTA TSFMs, we conclude two insights regarding the current progress and direction of the field.

\subsection{Architectural Progress}
The scaling curves of our generic Patch Transformer show performance comparable to that of the recent SOTA TSFMs with similar sizes, matching other deliberately designed architectures. This suggests that for large-scale time series modeling and pretraining, the ``vanilla'' Transformer architecture remains a competitive choice by demonstrating high forecasting performance and scalability.
Alternatively, we also observe distinct advantages of State-Space-Models (TiRex and FlowState) which demonstrate better parameter efficiency, achieving competitive results with more compact models. Notably, FlowState achieves strong performance with \emph{only real data from GIFT-Eval-Pretrain}, suggesting that timestamp-level modeling via SSMs may offer a valuable direction for future research. However, the scalability of SSMs has not yet been demonstrated. 

\subsection{The Role of Data and Training Protocol}
In Figure \ref{fig:size-crps}, the vertical trajectory (purple line) shows the substantial impact of data composition on the quality and performance of TSFMs, which is further supported by Section \ref{sec:ablation-data} and Figure \ref{fig:gift-benchmark-no-leak}. 
Furthermore, our ablations in Section \ref{sec:ablation} identify that specific training techniques, especially CPM and extended context lengths, are essential mechanisms for learning dynamics of time series and performing generalized predictions. Additionally, as presented in Section \ref{sec:others}, we demonstrate that other generic, standard architectures could also achieve comparable performance as the deliberately designed TSFMs when coupled with our training protocol. These findings suggest that the performance leaps observed in recent studies may stem, at least in part, from advancements in training methodology rather than from innovation in the core architecture.

\subsection{Recommendations for Future Research}

Based on our findings, we make following recommendations for the community of TSFM research:
\begin{itemize}[nosep, leftmargin=0.3cm]
    \item \textbf{Benchmark Artifacts and Architectures Separately}: We should distinguish between releasing high-utility checkpoints (an engineering goal) and discovering new architectures (a scientific goal). Releasing models trained on specific data is valuable, but it often obscures the true source of performance gains. To measure scientific progress accurately, we advocate evaluating architectures independently of their pretraining data. Future research should explicitly state whether contributions stem from architectural innovation or data curation.
    
    \item \textbf{Standardize the Pretraining Corpus}: To enable fair comparisons, the community needs a unified, publicly available pretraining corpus, analogous to ImageNet. Currently, heterogeneous pretraining datasets make cross-paper architectural comparisons virtually impossible. With our proven protocol as a proposal and potential starting point, we advocate for a standardized framework where models are trained on a fixed, shared dataset to isolate architectural innovations from data curation techniques.
    
    \item \textbf{Bridging Research and Application with High-Quality Benchmark Datasets}: Standardized benchmarks must not compromise model quality. Our results show that data diversity and broad domain coverage have a substantial effect on performance. Therefore, a unified pretraining corpus must be large and diverse enough to train models near ``production-grade'' levels, rather than relying on limited ``toy'' datasets. A high-quality standard corpus ensures that architectural improvements observed in research will reliably translate to real-world applications.
\end{itemize}


\subsection{Conclusion}

In this work, we revisited the generic Patch Transformer to evaluate its effectiveness as a Time Series Foundation Model. We demonstrated that this generic architecture, when supported by an effective training recipe, achieves performance statistically indistinguishable from that of the leading TSFMs on the GIFT-Eval and TIME forecasting benchmarks. Through extensive ablation studies, we examined the factors contributing to this success, finding that pretraining data composition, contiguous patch masking, and context length have the largest effect within our protocol.

These findings imply that the rapid leaderboard advancements observed in recent TSFMs may be partially confounded by non-uniform training protocols, which obscure the actual progress in architectural innovations. By establishing this robust baseline, we highlight the central role of data and training techniques in current benchmarks. We hope this work encourages the community to adopt more rigorous benchmarking protocols, redirecting focus toward architectural innovations that demonstrate clear value beyond data engineering.

\begin{ack}
This work was supported by IBM through the IBM-Rensselaer Future of Computing Research Collaboration (FCRC).
The authors acknowledge the National Artificial Intelligence Research Resource (NAIRR) Pilot and Mass Open Cloud for providing computational resources that contributed to this work.
\end{ack}

\newpage
\bibliographystyle{apalike}
\bibliography{reference}

@article{ansari2025chronos2,
  title={Chronos-2: From univariate to universal forecasting},
  author={Ansari, Abdul Fatir and Shchur, Oleksandr and K{\"u}ken, Jaris and Auer, Andreas and Han, Boran and Mercado, Pedro and Rangapuram, Syama Sundar and Shen, Huibin and Stella, Lorenzo and Zhang, Xiyuan and others},
  journal={arXiv preprint arXiv:2510.15821},
  year={2025}
}

@inproceedings{deng2009imagenet,
  author={Deng, Jia and Dong, Wei and Socher, Richard and Li, Li-Jia and Kai Li and Li Fei-Fei},
  booktitle={2009 IEEE Conference on Computer Vision and Pattern Recognition}, 
  title={ImageNet: A large-scale hierarchical image database}, 
  year={2009},
  volume={},
  number={},
  pages={248-255},
}

@inproceedings{ekambaram2023mlpmixer,
  title={Tsmixer: Lightweight mlp-mixer model for multivariate time series forecasting},
  author={Ekambaram, Vijay and Jati, Arindam and Nguyen, Nam and Sinthong, Phanwadee and Kalagnanam, Jayant},
  booktitle={Proceedings of the 29th ACM SIGKDD conference on knowledge discovery and data mining},
  pages={459--469},
  year={2023}
}

@article{dao2024mamba,
  title={Transformers are ssms: Generalized models and efficient algorithms through structured state space duality},
  author={Dao, Tri and Gu, Albert},
  journal={arXiv preprint arXiv:2405.21060},
  year={2024}
}

@article{
ansari2024chronos,
title={Chronos: Learning the Language of Time Series},
author={Abdul Fatir Ansari and Lorenzo Stella and Ali Caner Turkmen and Xiyuan Zhang and Pedro Mercado and Huibin Shen and Oleksandr Shchur and Syama Sundar Rangapuram and Sebastian Pineda Arango and Shubham Kapoor and Jasper Zschiegner and Danielle C. Maddix and Hao Wang and Michael W. Mahoney and Kari Torkkola and Andrew Gordon Wilson and Michael Bohlke-Schneider and Bernie Wang},
journal={Transactions on Machine Learning Research},
issn={2835-8856},
year={2024},
}

@article{gloeckle2024mtpllm,
  title={Better \& faster large language models via multi-token prediction},
  author={Gloeckle, Fabian and Idrissi, Badr Youbi and Rozi{\`e}re, Baptiste and Lopez-Paz, David and Synnaeve, Gabriel},
  journal={arXiv preprint arXiv:2404.19737},
  year={2024}
}

@inproceedings{hafner2019rssm,
  title={Learning latent dynamics for planning from pixels},
  author={Hafner, Danijar and Lillicrap, Timothy and Fischer, Ian and Villegas, Ruben and Ha, David and Lee, Honglak and Davidson, James},
  booktitle={International conference on machine learning},
  pages={2555--2565},
  year={2019},
  organization={PMLR}
}

@article{qiao2026time,
  title={It's TIME: Towards the Next Generation of Time Series Forecasting Benchmarks},
  author={Qiao, Zhongzheng and Pan, Sheng and Wang, Anni and Zhukova, Viktoriya and Liu, Yong and Jiang, Xudong and Wen, Qingsong and Long, Mingsheng and Jin, Ming and Liu, Chenghao},
  journal={arXiv preprint arXiv:2602.12147},
  year={2026}
}

@article{liu2026timers1,
  title={Timer-S1: A Billion-Scale Time Series Foundation Model with Serial Scaling},
  author={Liu, Yong and Su, Xingjian and Wang, Shiyu and Zhang, Haoran and Liu, Haixuan and Wang, Yuxuan and Ye, Zhou and Xiang, Yang and Wang, Jianmin and Long, Mingsheng},
  journal={arXiv preprint arXiv:2603.04791},
  year={2026}
}

@inproceedings{
auer2025tirex,
title={TiRex: Zero-Shot Forecasting Across Long and Short Horizons with Enhanced In-Context Learning},
author={Andreas Auer and Patrick Podest and Daniel Klotz and Sebastian B{\"o}ck and G{\"u}nter Klambauer and Sepp Hochreiter},
booktitle={The Thirty-ninth Annual Conference on Neural Information Processing Systems},
year={2025},
}

@misc{
aksu2025gifteval,
title={{GIFT}-Eval: A Benchmark for General Time Series Forecasting Model Evaluation},
author={Taha Aksu and Gerald Woo and Juncheng Liu and Xu Liu and Chenghao Liu and Silvio Savarese and Caiming Xiong and Doyen Sahoo},
year={2025},
}

@inproceedings{
woo2024moirai,
title={Unified Training of Universal Time Series Forecasting Transformers},
author={Gerald Woo and Chenghao Liu and Akshat Kumar and Caiming Xiong and Silvio Savarese and Doyen Sahoo},
booktitle={Forty-first International Conference on Machine Learning},
year={2024},
}

@inproceedings{
das2024timesfm,
title={A decoder-only foundation model for time-series forecasting},
author={Abhimanyu Das and Weihao Kong and Rajat Sen and Yichen Zhou},
booktitle={Forty-first International Conference on Machine Learning},
year={2024},
}

@inproceedings{
liu2025sundial,
title={Sundial: A Family of Highly Capable Time Series Foundation Models},
author={Yong Liu and Guo Qin and Zhiyuan Shi and Zhi Chen and Caiyin Yang and Xiangdong Huang and Jianmin Wang and Mingsheng Long},
booktitle={Forty-second International Conference on Machine Learning},
year={2025},
}

@inproceedings{
hoo2024tabpfnts,
title={The Tabular Foundation Model Tab{PFN} Outperforms Specialized Time Series Forecasting Models Based on Simple Features},
author={Shi Bin Hoo and Samuel M{\"u}ller and David Salinas and Frank Hutter},
booktitle={NeurIPS 2024 Third Table Representation Learning Workshop},
year={2024},
}

@inproceedings{
chen2025visionts,
title={Vision{TS}: Visual Masked Autoencoders Are Free-Lunch Zero-Shot Time Series Forecasters},
author={Mouxiang Chen and Lefei Shen and Zhuo Li and Xiaoyun Joy Wang and Jianling Sun and Chenghao Liu},
booktitle={Forty-second International Conference on Machine Learning},
year={2025},
}

@inproceedings{
cohen2025toto,
title={This Time is Different: An Observability Perspective on Time Series Foundation Models},
author={Ben Cohen and Emaad Khwaja and Youssef Doubli and Salahidine Lemaachi and Chris Lettieri and Charles Masson and Hugo Miccinilli and Elise Ram{\'e} and Qiqi Ren and Afshin Rostamizadeh and Jean Ogier du Terrail and Anna-Monica Toon and Kan Wang and Stephan Xie and Zongzhe Xu and Viktoriya Zhukova and David Asker and Ameet Talwalkar and Othmane Abou-Amal},
booktitle={The Thirty-ninth Annual Conference on Neural Information Processing Systems},
year={2025},
}

@inproceedings{
graf2025flowstate,
title={FlowState: Sampling-Rate\nobreakdash-Equivariant Time\nobreakdash-Series Forecasting},
author={Lars Graf and Thomas Ortner and Stanis{\l}aw Wo{\'z}niak and Angeliki Pantazi},
booktitle={Forty-third International Conference on Machine Learning},
year={2026},
}

@article{wang2025yinglong,
  title={Output Scaling: YingLong-Delayed Chain of Thought in a Large Pretrained Time Series Forecasting Model},
  author={Wang, Xue and Zhou, Tian and Gao, Jinyang and Ding, Bolin and Zhou, Jingren},
  journal={arXiv preprint arXiv:2506.11029},
  year={2025}
}

@inproceedings{nie2023patchtst,
  title     = {A Time Series is Worth 64 Words: Long-term Forecasting with Transformers},
  author    = {Nie, Yuqi and
               H. Nguyen, Nam and
               Sinthong, Phanwadee and 
               Kalagnanam, Jayant},
  booktitle = {International Conference on Learning Representations},
  year      = {2023}
}

@inproceedings{vaswani2017attention,
	title = {Attention is All you Need},
	volume = {30},
	booktitle = {Advances in {Neural} {Information} {Processing} {Systems}},
	author = {Vaswani, Ashish and Shazeer, Noam and Parmar, Niki and Uszkoreit, Jakob and Jones, Llion and Gomez, Aidan N and Kaiser, Lukasz and Polosukhin, Illia},
	year = {2017},
}

@inproceedings{
beck2024xlstm,
title={x{LSTM}: Extended Long Short-Term Memory},
author={Maximilian Beck and Korbinian P{\"o}ppel and Markus Spanring and Andreas Auer and Oleksandra Prudnikova and Michael K Kopp and G{\"u}nter Klambauer and Johannes Brandstetter and Sepp Hochreiter},
booktitle={The Thirty-eighth Annual Conference on Neural Information Processing Systems},
year={2024},
}

@inproceedings{
smith2023ssm,
title={Simplified State Space Layers for Sequence Modeling},
author={Jimmy T.H. Smith and Andrew Warrington and Scott Linderman},
booktitle={The Eleventh International Conference on Learning Representations },
year={2023},
}

@inproceedings{
kim2022revin,
title={Reversible Instance Normalization for Accurate Time-Series Forecasting against Distribution Shift},
author={Taesung Kim and Jinhee Kim and Yunwon Tae and Cheonbok Park and Jang-Ho Choi and Jaegul Choo},
booktitle={International Conference on Learning Representations},
year={2022}
}

@inproceedings{he2022mae,
  title={Masked autoencoders are scalable vision learners},
  author={He, Kaiming and Chen, Xinlei and Xie, Saining and Li, Yanghao and Doll{\'a}r, Piotr and Girshick, Ross},
  booktitle={Proceedings of the IEEE/CVF conference on computer vision and pattern recognition},
  pages={16000--16009},
  year={2022}
}

@inproceedings{
    wen2024abstracted,
    title={Abstracted Shapes as Tokens - A Generalizable and Interpretable Model for Time-series Classification},
    author={Yunshi Wen and Tengfei Ma and Tsui-Wei Weng and Lam M. Nguyen and Anak Agung Julius},
    booktitle={The Thirty-eighth Annual Conference on Neural Information Processing Systems},
    year={2024}
}

@inproceedings{goswami2024moment,
  title={MOMENT: A Family of Open Time-series Foundation Models},
  author={Mononito Goswami and Konrad Szafer and Arjun Choudhry and Yifu Cai and Shuo Li and Artur Dubrawski},
  booktitle={International Conference on Machine Learning},
  year={2024}
}

@inproceedings{
    shi2025timemoe,
    title={Time-MoE: Billion-Scale Time Series Foundation Models with Mixture of Experts},
    author={Xiaoming Shi and Shiyu Wang and Yuqi Nie and Dianqi Li and Zhou Ye and Qingsong Wen and Ming Jin},
    booktitle={The Thirteenth International Conference on Learning Representations},
    year={2025}
}

@inproceedings{
    godahewa2021monash,
    title={Monash Time Series Forecasting Archive},
    author={Rakshitha Wathsadini Godahewa and Christoph Bergmeir and Geoffrey I. Webb and Rob Hyndman and Pablo Montero-Manso},
    booktitle={Thirty-fifth Conference on Neural Information Processing Systems Datasets and Benchmarks Track (Round 2)},
    year={2021}
}

@article{jiang2023libcity,
  title={Libcity: A unified library towards efficient and comprehensive urban spatial-temporal prediction},
  author={Jiang, Jiawei and Han, Chengkai and Jiang, Wenjun and Zhao, Wayne Xin and Wang, Jingyuan},
  journal={arXiv preprint arXiv:2304.14343},
  year={2023}
}

@inproceedings{
    emami2023buildingsbench,
    title={BuildingsBench: A Large-Scale Dataset of 900K Buildings and Benchmark for Short-Term Load Forecasting},
    author={Patrick Emami and Abhijeet Sahu and Peter Graf},
    booktitle={Thirty-seventh Conference on Neural Information Processing Systems Datasets and Benchmarks Track},
    year={2023}
}

@inproceedings{
nguyen2023climatelearn,
title={ClimateLearn: Benchmarking Machine Learning for Weather and Climate Modeling},
author={Tung Nguyen and Jason Kyle Jewik and Hritik Bansal and Prakhar Sharma and Aditya Grover},
booktitle={Thirty-seventh Conference on Neural Information Processing Systems Datasets and Benchmarks Track},
year={2023}
}

@inproceedings{
liu2023largest,
title={Large{ST}: A Benchmark Dataset for Large-Scale Traffic Forecasting},
author={Xu Liu and Yutong Xia and Yuxuan Liang and Junfeng Hu and Yiwei Wang and Lei Bai and Chao Huang and Zhenguang Liu and Bryan Hooi and Roger Zimmermann},
booktitle={Thirty-seventh Conference on Neural Information Processing Systems Datasets and Benchmarks Track},
year={2023}
}

@inproceedings{wu2021autoformer,
 author = {Wu, Haixu and Xu, Jiehui and Wang, Jianmin and Long, Mingsheng},
 booktitle = {Advances in Neural Information Processing Systems},
 title = {Autoformer: Decomposition Transformers with Auto-Correlation for Long-Term Series Forecasting},
 year = {2021}
}

@inproceedings{zhou2021informer,
  title={Informer: Beyond efficient transformer for long sequence time-series forecasting},
  author={Zhou, Haoyi and Zhang, Shanghang and Peng, Jieqi and Zhang, Shuai and Li, Jianxin and Xiong, Hui and Zhang, Wancai},
  booktitle={Proceedings of the AAAI conference on artificial intelligence},
  volume={35},
  pages={11106--11115},
  year={2021}
}

@inproceedings{palaskar2024automixer,
  title={Automixer for improved multivariate time-series forecasting on business and it observability data},
  author={Palaskar, Santosh and Ekambaram, Vijay and Jati, Arindam and Gantayat, Neelamadhav and Saha, Avirup and Nagar, Seema and Nguyen, Nam H and Dayama, Pankaj and Sindhgatta, Renuka and Mohapatra, Prateeti and others},
  booktitle={Proceedings of the AAAI conference on artificial intelligence},
  volume={38},
  pages={22962--22968},
  year={2024}
}

@inproceedings{shen2015statistical,
  title={Statistical characterization of business-critical workloads hosted in cloud datacenters},
  author={Shen, Siqi and Van Beek, Vincent and Iosup, Alexandru},
  booktitle={2015 15th IEEE/ACM international symposium on cluster, cloud and grid computing},
  pages={465--474},
  year={2015},
  organization={IEEE}
}

@misc{howard2017restaurant,
    author = {Addison Howard and Haruka Yui and Mark McDonald and Will Cukierski},
    title = {Recruit Restaurant Visitor Forecasting},
    year = {2017},
    howpublished = {\url{https://kaggle.com/competitions/recruit-restaurant-visitor-forecasting}},
    note = {Kaggle}
}

@inproceedings{lai2018modeling,
  title={Modeling long-and short-term temporal patterns with deep neural networks},
  author={Lai, Guokun and Chang, Wei-Cheng and Yang, Yiming and Liu, Hanxiao},
  booktitle={The 41st international ACM SIGIR conference on research \& development in information retrieval},
  pages={95--104},
  year={2018}
}

@misc{trindade2015electricity,
  author       = {Trindade, Artur},
  title        = {{ElectricityLoadDiagrams20112014}},
  year         = {2015},
  howpublished = {UCI Machine Learning Repository},
  note         = {{DOI}: https://doi.org/10.24432/C58C86}
}

@inproceedings{devlin2019bert,
  title={Bert: Pre-training of deep bidirectional transformers for language understanding},
  author={Devlin, Jacob and Chang, Ming-Wei and Lee, Kenton and Toutanova, Kristina},
  booktitle={Proceedings of the 2019 conference of the North American chapter of the association for computational linguistics: human language technologies, volume 1 (long and short papers)},
  pages={4171--4186},
  year={2019}
}


\appendix

\section{Related Work}

To consume large-scale pretraining data, TSFMs adopt scalable architectures paired with explicit input processing or ``tokenization'' strategies. 
Following its success in computer vision and natural language processing, the Transformer architecture \citep{vaswani2017attention} also serves as the dominant backbone choice for TSFMs. 
For time series data, tokenization approaches can vary between timestamp-level and patch-level granularities.

For instance, Chronos \citep{ansari2024chronos} discretizes values at the timestamp level to construct a ``language of time series,'' enabling forecasting via a decoder-only autoregressive Transformer.
TimesFM \citep{das2024timesfm} and Moirai \citep{woo2024moirai} adopt patch-level tokenization to capture local semantics and improve computational efficiency, utilizing decoder-only and encoder-only (i.e., full-attention Transformer with masked inputs) architectures, respectively.
Very recently, Chronos-2 \citep{ansari2025chronos2} extends the field from univariate TSFM to generic multivariate modeling and forecasting, while heavily relying on synthetic data.
Beyond deterministic mapping, Sundial \citep{liu2025sundial} integrates patch-level modeling with flow-matching decoding to provide more ``natural'' probabilistic forecasts from point predictions. 

Recent works also explore state-space models as alternatives to the Transformer backbone, achieving high performance with relatively compact models. For example, TiRex \citep{auer2025tirex} combines patch-level processing with the xLSTM \citep{beck2024xlstm} architecture. FlowState \citep{graf2025flowstate} formulates timestamp-level inputs as a discrete ODE using an S5 backbone \citep{smith2023ssm}, utilizing a specialized scaling factor to handle various sampling rates. 

However, while these models continue to rapidly advance the SOTA results, their heterogeneous training protocols obscure the attribution of these gains. In particular, the increasing reliance on proprietary synthetic data generation pipelines, where the exact synthetic corpus is often not publicly released, introduces a significant uncontrolled variable. Consequently, it remains unclear whether recent improvements stem from the core architectural innovations or from superior data engineering (see Appendix \ref{append:baselines} for details).

\section{Broader Impacts}

This work aims to support the development of TSFMs by establishing a transparent and reproducible baseline. We discuss the broader impacts of our contributions along three dimensions.

\paragraph{Establishing Transparent Baselines.}
Many existing TSFM studies emphasize final model performance while remaining opaque about their training protocols. Our results show that the protocol can be as essential as the architecture itself in determining downstream performance. By releasing our model checkpoints, documenting our training protocol in full, and reporting detailed empirical findings, we provide a strong and reproducible baseline that future work can build upon. We hope this encourages a research culture in which architectural contributions are reported alongside the training conditions under which they were obtained, so that genuine progress in architecture can be more reliably identified.

\paragraph{Advancing a Unified Pretraining Corpus.}
Fair benchmarking of TSFM architectures requires a high-quality and standardized pretraining corpus. Our pretraining data composition is openly documented and can be reproduced from public sources, which we view as an initial step toward such a standard. We hope this effort motivates the community to develop even larger and more diverse corpora in the future. Standardized evaluation benchmarks alone are not sufficient. A shared pretraining corpus is equally important for isolating the contribution of a new architecture from the contribution of its training data, analogous to the role that standardized benchmarks have played in accelerating progress in adjacent fields such as computer vision \citep{deng2009imagenet}.

\paragraph{Lowering the Barrier to Entry.}
Developing a competitive TSFM currently requires substantial resources and considerable trial and error to identify an effective training recipe. This creates a high barrier for entry-level researchers and small research groups, and it makes it difficult to determine whether a new architecture is genuinely ineffective or simply trained under a suboptimal protocol. By providing a protocol that produces a generic Transformer competitive with models from leading industrial groups, we hope to lower this barrier and enable a broader range of researchers to contribute to the field.

\section{Implementation Details} \label{append:model-details}

\subsection{Mask-Aware Normalization} \label{sec:normalization}

Heterogeneous value distribution and non-stationarity are major challenges in time-series modeling, where Reversible Instance Normalization \citep{kim2022revin} is proposed to mitigate such issues. In our masked autoencoding formulation, computing statistics including the masked (target) values would introduce information leakage about the target distribution.

To address this, we utilize Mask-Aware Normalization, which computes the mean $\mu$ and standard deviation $\sigma$ using only the visible time points:
\begin{equation}
    \mu = \frac{\sum_{t=1}^{T} \mathds{1}(m_t=0)\, x_t}{\sum_{t=1}^{T} \mathds{1}(m_t=0)},
\qquad
\sigma = \sqrt{\frac{\sum_{t=1}^{T} \mathds{1}(m_t=0)\, (x_t - \mu)^2}{\sum_{t=1}^{T} \mathds{1}(m_t=0)}}.
\end{equation}
where $\mathds{1}(\cdot)$ denotes the indicator function. These statistics are stored and applied later for Reverse Normalization to restore the predicted quantiles into the original scale of the input data. Furthermore, we follow \citet{ansari2025chronos2} to normalize the input time series with
$\bar{\mathbf{x}} = \texttt{sinh}^{-1} \left( \frac{\mathbf{x} - \mu}{\sigma} \right)$, which suppresses the influence of extreme outliers.

\subsection{Masking Strategy} \label{sec:masking_details}

We provide supplementary details regarding the input processing described in Section \ref{sec:architecture}. The design of the input mask is critical for ensuring model flexibility and generalized usage. 

\paragraph{Mask Processing} In our implementation, the unified input mask $\mathbf{m}$ is a composite of three distinct components:
\begin{itemize}
    \item Prediction mask $\mathbf{m}^{\text{pred}}$: Designates the observed values in the sequence that are dropped to serve as prediction targets. To optimize the model for forecasting tasks, with probability $0.5$ we additionally mask a contiguous block of up to $4 \cdot N_{\text{CPM}}$ patches at the end of each training sample.
    \item Missing mask $\mathbf{m}^{\text{miss}}$: Identifies inherent missing values within the raw data. These values are retained in the attention mechanism (to preserve temporal position) but are marked as missing in $\mathbf{m}$ and strictly excluded from the loss calculation.
    \item Padding mask $\mathbf{m}^{\text{pad}}$: Identifies padding tokens used to fill the context window when the input sequence length is shorter than the context window $T$. We apply left-padding to align the sequence. Patches consisting entirely of padding values are masked out of the attention mechanism, where we empirically find this approach could improve performance by saving attention.
\end{itemize}
The model processes the logical union of these masks, defined as $\mathbf{m} = \mathbf{m}^{\text{pred}} \vee \mathbf{m}^{\text{miss}} \vee \mathbf{m}^{\text{pad}}$. Note that in Section \ref{sec:architecture} and Equation \ref{eq:loss}, we simplified the notation to assume $\mathbf{m} \approx \mathbf{m}^{\text{pred}}$ for clarity of presentation.

\paragraph{Adaptive Contiguous Patch Masking} CPM masks blocks of $N_{\text{CPM}} \cdot L$ consecutive observations. However, applying a fixed block size to short input sequences can inadvertently mask the entire valid history. Therefore, we implement an Adaptive CPM strategy. Formally for a time series $\mathbf{x} \in \mathbb{R}^{T^{(x)}}$ with length $T^{(x)} \ll T$, we dynamically adjust the block size $N_{\text{CPM}}^{(x)}$ as follows:
\begin{equation*}
   N_{\text{CPM}}^{(x)} = \min \left(\left\lceil \frac{\lceil T^{(x)} / L \rceil }{4}  \right\rceil , N_{\text{CPM}} \right).
\end{equation*}
This preserves sufficient context for the model to learn meaningful dependencies even in short samples.

\subsection{Quantile Head and Inference Pipeline} \label{sec:quantile_head}

\paragraph{Pipeline} Given a raw series, we compute the normalization statistics $\mu$ and $\sigma$ from the visible context only (Appendix \ref{sec:normalization}), apply the $\sinh^{-1}$ transform, segment the series into non-overlapping patches of length $L$, and mark the forecast positions as masked. The encoder processes the visible and masked patches jointly. The quantile head, implemented as a residual block, maps the embedding of each masked patch to $K=99$ quantile values at levels $\{0.01, 0.02, \dots, 0.99\}$ for every timestamp in the patch. During training, the pinball loss in Equation \ref{eq:loss} is computed only on masked positions that correspond to real observations; missing and padded positions are excluded. At inference, all target patches are decoded in a single non-autoregressive forward pass, and the predicted quantiles are mapped back to the original scale with $\mathbf{y} = \mathtt{sinh}(\hat{\mathbf{q}}) \cdot \sigma + \mu$.

\paragraph{Monotone Quantiles} The head does not predict the $K$ quantile values independently. For each timestamp $t$, it predicts an anchor value $b_t$ and $K$ raw increments $\delta_{t,1}, \dots, \delta_{t,K}$, and outputs
\begin{equation}
    \hat{q}_t^{(\tau_k)} = b_t + \sum_{j=1}^{k} \frac{\operatorname{softplus}(\delta_{t,j})}{K}, \qquad k = 1, \dots, K,
\end{equation}
so that $\hat{q}_t^{(\tau_k)} - \hat{q}_t^{(\tau_{k-1})} = \operatorname{softplus}(\delta_{t,k})/K > 0$ for $k = 2, \dots, K$. The predicted quantiles are therefore ordered by construction, and no post-hoc sorting is required to prevent quantile crossing.

\paragraph{Non-overlapping Patches} We use non-overlapping patches (stride equal to $L$) throughout. This keeps the token count at $T/L$ and avoids redundant computation over overlapping windows, at the possible cost of fine-scale information at patch boundaries. Patch overlap was not ablated in this study; a patch-size ablation is reported in Appendix \ref{append:patch-size}.

\newpage
\subsection{Hyperparameters}

\begin{table}[h]
    \centering
    \caption{Detailed Hyperparameter Configuration. The values are selected based on standard practices in TSFMs and Transformers.}
    \label{tab:hyperparameters}
    \begin{tabular}{l l p{4.7cm}}
        \toprule
        \textbf{Hyperparameter} & \textbf{Value} & \textbf{Rationale / Source} \\
        \midrule
        \rowcolor{gray!15} \multicolumn{3}{l}{\textbf{Architecture}} \\
        Context Length ($T$) & 8192 & A large context window for general usage. \citep{ansari2025chronos2} \\
        Patch Size ($L$) & 16 & Non-overlapping (stride $=L$), fixed across datasets and frequencies, following PatchTST \citep{nie2023patchtst}; see Appendix \ref{append:patch-size}. \\
        Number of Layers ($N_{\text{layer}}$) & 20 & Modified from ViT-Large. \\
        Embedding Dimension ($d$) & 1024 & From ViT-Large. \\
        Attention Head Dimension & 64 & Standard choice for Transformers. \\
        Number of Attention Heads & $d/64$ & Standard choice for Transformers. \\
        Number of Quantiles ($K$) & 99 &  \\
        Quantile Levels $Q$ & $\{0.01, 0.02, \dots, 0.99\}$ & \\
        \midrule
        \rowcolor{gray!15} \multicolumn{3}{l}{\textbf{Training Configurations}} \\
        Masking Ratio & 40\% & An intermediate value based on BERT \citep{devlin2019bert} and MAE \citep{he2022mae}. \\
        CPM Block Size ($N_{CPM}$) & 8 &  \\
        
        \midrule
        \rowcolor{gray!15} \multicolumn{3}{l}{\textbf{Optimization}} \\
        Total Training Steps & 100k &  \\
        Effective Batch Size & 2048 \\
        Optimizer & AdamW & Standard choice for foundation models. \\
        Learning Rate Schedule & Cosine with Linear Warmup & Standard choice for foundation models. \\
        Peak Learning Rate & $3 \cdot 10^{-4}$ & Standard choice, conservative for training stability.\\
        Minimum Learning Rate & $1 \cdot 10^{-5}$ \\
        Weight Decay & 0.1 & Empirical choice to ensure training stability. \\
        $(\beta_1, \beta_2)$ in AdamW & $(0.9, .95)$ & Empirical choice to ensure training stability. \\
        Gradient Norm Clipping & 1 & Standard choice, conservative for training stability.\\
        Precision & FP16 Mixed & Accelerate training, supported by V100 GPUs. \\
        \bottomrule
    \end{tabular}
\end{table}

\newpage

\subsection{Alternate Generic Architectures}

In this section, we detail the formulations of the three additional generic architectures discussed in Section \ref{sec:others} and Table \ref{tab:gift-architecture}. All three share the same normalization and patching pipeline as the Transformer backbone, differing only in how the resulting patch embeddings are processed. We deliberately instantiate each backbone in its most canonical form so that it cleanly represents a single type of architectural inductive bias, enabling a controlled comparison of their relative effectiveness. In practice, these primitives are often combined to achieve broader and more robust utility (e.g., SSM+MLP, CNN+MLP hybrids).

\paragraph{MLP-Mixer backbone.}
The backbone consists of $L$ identical Mixer blocks operating on a univariate token sequence $\mathbf{X} \in \mathbb{R}^{N \times d_{\text{model}}}$, where $N$ is the number of patches and each token is the embedding of one patch. Each block alternates two MLPs that act along orthogonal axes of the token matrix: an \emph{inter-patch} MLP applied across the patch dimension and an \emph{intra-patch} MLP applied across the embedding dimension. Both sub-layers follow a pre-norm residual formulation:
\begin{align}
    \mathbf{U}_\ell &= \mathbf{X}_\ell + \big[ f_{\text{inter}}\!\left( \text{LN}(\mathbf{X}_\ell)^\top \right) \big]^\top, \\
    \mathbf{X}_{\ell+1} &= \mathbf{U}_\ell + f_{\text{intra}}\!\left( \text{LN}(\mathbf{U}_\ell) \right),
\end{align}
where $f_{\text{inter}}$ and $f_{\text{intra}}$ are two-layer MLPs with GELU activation and dropout. Transposition before and after $f_{\text{inter}}$ exposes the patch axis as the input dimension of the MLP, allowing it to learn position-dependent mixing weights over the entire sequence in a single operation.

\paragraph{Bidirectional Mamba2 backbone.}
The backbone is a stack of $L$ bidirectional Mamba-2 blocks operating on a token sequence $\mathbf{X} \in \mathbb{R}^{N \times d_{\text{model}}}$. Each block applies two independent Mamba-2 selective state-space modules \citep{dao2024mamba} to opposite traversal orders of the same layer-normalized input and combines their outputs additively within a single residual connection:
\begin{equation}
    \mathbf{X}_{\ell+1} = \mathbf{X}_\ell + \text{SSM}_{\rightarrow}\!\left(\text{LN}(\mathbf{X}_\ell)\right) + \mathcal{R}\!\left(\text{SSM}_{\leftarrow}\!\left(\mathcal{R}(\text{LN}(\mathbf{X}_\ell))\right)\right),
\end{equation}
where $\mathcal{R}$ denotes reversal along the patch axis and $\text{SSM}_{\rightarrow}, \text{SSM}_{\leftarrow}$ are two separately parameterized Mamba-2 modules. Summing rather than concatenating the two directional outputs preserves the channel dimension and avoids doubling the parameter count of downstream projections.

This bidirectional construction is motivated by the non-causal nature of the main modeling setup with the encoder-only Transformer: at inference time the entire context window is observed, so restricting the backbone to left-to-right state propagation would discard information that a unidirectional SSM cannot recover.

\paragraph{CNN-ResNet backbone.} 
The backbone is a stack of $L$ residual convolutional blocks operating on a token sequence $\mathbf{X} \in \mathbb{R}^{d_{\text{model}} \times N}$, where $N$ is the number of patches. Each block applies two 1D convolutions with "same" (non-causal) padding, separated by a GELU activation and dropout, followed by a residual connection and a LayerNorm along the channel dimension:
\begin{equation}
    \mathbf{X}_{\ell+1} = \text{LN}\!\left( \mathbf{X}_\ell + \text{Conv}^{(2)}_{k_\ell}\!\left(\sigma\!\left(\text{Conv}^{(1)}_{k_\ell}(\mathbf{X}_\ell)\right)\right) \right),
\end{equation}
where $\sigma$ denotes GELU and $k_\ell$ is the kernel size of the $\ell$-th block. All convolutions preserve sequence length, so the temporal resolution is unchanged throughout the network.

To capture dependencies at multiple temporal scales, we adopt a coarse-to-fine kernel schedule that monotonically decreases with depth, with the default configuration $\{31, 21, 15, 11, 7, 3\}$ for $L = 6$. Early layers employ large kernels to aggregate broad context and expose low-frequency structure, while later layers use progressively smaller kernels to refine local, high-frequency patterns.

\subsection{Experiment Framework} \label{append:framework}
Our models, training pipeline, and experiments are implemented using PyTorch and the Hugging Face ecosystem. To ensure broad compatibility with open-source frameworks, we leverage standard libraries including the \texttt{PretrainedModel} class, \texttt{Trainer}, \texttt{Accelerate}, and \texttt{Datasets}. Unless noted below, pretraining was conducted on NVIDIA V100 GPUs using FP16 mixed-precision training to optimize computational efficiency.

\paragraph{Compute} All pretraining runs reported in this paper, including every ablation variant, use the same training configuration (Table \ref{tab:hyperparameters}). With the exception noted below, they were run on 96 NVIDIA V100-32G GPUs for approximately 13 hours per run, i.e., 1,248 allocated V100 GPU-hours. The 96-GPU allocation was chosen for its host memory, which keeps the full corpus in memory, rather than for model parallelism. For reference, we reproduced the full-scale run on 8 NVIDIA H100-80G GPUs in about 18 hours (144 H100 GPU-hours), and in about 6.5 hours (52 H100 GPU-hours) with FlashAttention and by skipping computation on padding tokens; these H100 runs serve only as timing references, and none of the reported results uses their checkpoints. Owing to a change in the compute available to us, the patch-size ablation in Appendix \ref{append:patch-size} was run later on 32 NVIDIA A100-80G GPUs with the same training configuration.

\paragraph{Data Sampling} The pretraining corpus is organized as a collection of datasets: every dataset--frequency pair of GIFT-Eval-Pretrain, where each variate of a multivariate series is treated as a separate univariate series, together with the KernelSynth and TSMixup collections. Training samples are generated on the fly by the data loader in four steps, summarized in Algorithm \ref{alg:sampling}:
\begin{itemize}[nosep, leftmargin=0.3cm]
    \item \textbf{Series selection.} Datasets are sampled with equal probability, except that small datasets with fewer than 100 series are down-weighted by a factor of 100; within a dataset, series are drawn uniformly. Each series is standardized by its own mean and standard deviation.
    \item \textbf{Frequency augmentation.} For real series with a known sampling frequency, the original resolution is kept with probability at least $1/2$; otherwise the series is mean-pooled to one of up to three coarser calendar frequencies (e.g., hourly to daily), which widens the range of seasonal periods seen during pretraining.
    \item \textbf{Window cropping.} A window is cut at a random position. With probability $0.2$ its length is the full series length (capped at $T=8192$); otherwise it is drawn uniformly between $\ell_{\min}=64$ and the capped length.
    \item \textbf{Masking and padding.} Missing values are recorded in the missing mask, the adaptive CPM mask of Appendix \ref{sec:masking_details} is generated at the patch level, and the window and its masks are left-padded to length $T$.
\end{itemize}

\begin{algorithm}[h]
\caption{Generation of one pretraining sample}
\label{alg:sampling}
\begin{algorithmic}[1]
\Require Corpus $\mathcal{D}$ with dataset weights $\{w_d\}$; context length $T$; patch size $L$; masking ratio $\rho$; CPM block size $N_{\text{CPM}}$; minimum window length $\ell_{\min}$
\State Sample a dataset $d \propto w_d$ and a series $\mathbf{x} \in d$ uniformly; standardize $\mathbf{x}$ \Comment{series selection}
\State
\If{the sampling frequency of $d$ is known}
    \State Sample a pooling factor $s \in \{1, 1, 1, s_1, s_2, s_3\}$ and set $\mathbf{x} \leftarrow \mathrm{MeanPool}(\mathbf{x}, s)$ \Comment{frequency augmentation}
\EndIf
\State
\State With probability $0.2$ set $\ell \leftarrow \min(|\mathbf{x}|, T)$; otherwise sample $\ell \sim \mathcal{U}\{\ell_{\min}, \dots, \min(|\mathbf{x}|, T)\}$
\State
\State Sample a start $t_0 \sim \mathcal{U}\{0, \dots, |\mathbf{x}| - \ell\}$ and crop $\mathbf{x} \leftarrow \mathbf{x}_{t_0 : t_0 + \ell}$ \Comment{window cropping}
\State $\mathbf{m}^{\text{miss}} \leftarrow \mathrm{isnan}(\mathbf{x})$; replace missing values by $0$
\State $N \leftarrow \lceil \ell / L \rceil$;\; $N'_{\text{CPM}} \leftarrow \min(\lceil N/4 \rceil, N_{\text{CPM}})$ \Comment{adaptive CPM}
\State Sample $\lfloor \rho N / N'_{\text{CPM}} \rfloor$ block starts without replacement and mask $N'_{\text{CPM}}$ consecutive patches from each
\State With probability $0.5$, additionally mask up to $4 N'_{\text{CPM}}$ patches at the end of the window \Comment{forecast-style tail}
\State Expand the patch-level mask to $\mathbf{m}^{\text{pred}}$; left-pad $\mathbf{x}$, $\mathbf{m}^{\text{pred}}$, $\mathbf{m}^{\text{miss}}$ to length $T$ and record $\mathbf{m}^{\text{pad}}$
\State \Return $(\mathbf{x}, \mathbf{m}^{\text{pred}}, \mathbf{m}^{\text{miss}}, \mathbf{m}^{\text{pad}})$
\end{algorithmic}
\end{algorithm}

\newpage

\section{Benchmark and Baselines}

\subsection{GIFT-Eval} \label{append:gift-eval}

\paragraph{Evaluation Metrics} Here, we provide the detailed formulations of the GIFT-Eval metrics.

\textbf{MASE} is a metric for point forecasting. It compares the prediction's absolute error against the mean absolute error of a Naive forecasting baseline. Formally, given the target $\mathbf{x} \in \mathbb{R}^{T}$ and predicted median $\hat{\mathbf{q}}^{(0.5)} \in \mathbb{R}^{T}$, the MASE is calculated as follows:
\begin{equation}
    \mathtt{MASE}(\hat{\mathbf{q}}^{(0.5)}, \mathbf{x}) = \frac{\frac{1}{T} \sum_{t=1}^T \left| x_t - \hat{q}^{(0.5)} \right|}{\frac{1}{T-1} \sum_{t=2}^T \left| x_t - x_{t-1} \right|}.
\end{equation}

\textbf{CRPS} evaluates the accuracy of probabilistic forecasts. CRPS measures the difference between the predicted cumulative distribution functions (CDFs) and the target values. In practice, computing the CDFs via integration is computationally expensive. The GIFT-Eval benchmarking protocol approximates the CRPS using a discrete sum over a finite set of quantile levels, often referred to as the mean weighted quantile loss.

Let $\hat{\mathbf{q}} \in \mathbb{R}^{T \times K}$ denote the predicted values for the quantile levels $Q = \{\tau_1, \dots, \tau_K\}$. The approximated CRPS is defined as follows:
\begin{equation}
    \mathtt{CRPS}(\hat{\mathbf{q}}, \mathbf{x}) \approx \frac{1}{K} \sum_{k=1}^K \mathtt{wQL}[\tau_k], ~\text{where} ~~\mathtt{wQL}[\tau_k] = 2\frac{\sum_{t=1}^T (\tau_k - \mathds{1}(x_t < \hat{q}_t^{(\tau_k)}))(x_t - \hat{q}_t^{(\tau_k)})}{\sum_{t=1}^{T}|x_t|}.
\end{equation}

\paragraph{Dataset Information} The GIFT-Eval benchmark contains 55 datasets from 7 macro domains, including Nature, Web and Cloud Applications, Sales, Energy, Transportation, Economy and Finance, and Healthcare. The datasets are obtained from popular time series forecasting benchmarks, including Informer \citep{zhou2021informer}, Autoformer \citep{wu2021autoformer}, AutoMixer \citep{palaskar2024automixer}, LibCity \citep{jiang2023libcity}, LSTNet \citep{lai2018modeling}, Monash Archive \citep{godahewa2021monash}, Grid Workloads Archive \citep{shen2015statistical}, and Restaurant Competition \citep{howard2017restaurant}, and UCI ML Archive \citep{trindade2015electricity}. Please refer to \citet{aksu2025gifteval} for the detailed information about the GIFT-Eval and GIFT-Eval-Pretrain datasets.

\newpage
\subsection{Baselines} \label{append:baselines}

\begin{table}[ht]
\centering
\small 
\caption{Comparison of recent foundation models. Pretraining corpora sources are summarized for brevity. Publicly-available data are highlighted with \blue{blue}. Corpus overlap with the GIFT-Eval test cases are highlighted with \red{red}.}
\label{tab:tsfm-info}

\begin{tabularx}{\textwidth}{l c c r >{\RaggedRight\arraybackslash}X}
\toprule
\textbf{Model} & \textbf{Size} & \textbf{Arch.} & \textbf{Context Len.} & \textbf{Pretrain Corpus} \\
\midrule

TabPFN-TS \citep{hoo2024tabpfnts} & 11M & Encoder & $4096^{(1)}$ & Synthetic (Gaussian Process, Structured Causal Models). \\ \addlinespace

Sundial \citep{liu2025sundial} & 128M & Decoder & 2880 & \blue{Subset of Chronos}, \blue{LOTSA}, \blue{Synthetic}, Finance, IoT. \\ \addlinespace

YingLong \citep{wang2025yinglong} & 300M & Encoder & 8192 & \blue{Subset of Chronos}, \blue{KernelSynth}. \\ \addlinespace

Moirai-2 \citep{woo2024moirai} & 14M & Encoder & 512 & \blue{GIFT-Eval-Pretrain}, \red{Training split of GIFT-Eval}, \blue{TSMixup}, \blue{KernelSynth}, Internal Operations Data. Curated to remove the ``low-quality'' samples. \\ \addlinespace

Toto \citep{cohen2025toto} & 151M & Decoder & 4096 & Observability Metrics, \blue{GIFT-Eval-Pretrain}, \blue{Subset of Chronos}, Synthetic. \\ \addlinespace

FlowState \citep{graf2025flowstate} & 9.1M & S5 & 2048 & \blue{GIFT-Eval-Pretrain}. \\ \addlinespace

TimesFM-2.5 \citep{das2024timesfm} & 200M & Decoder & 16384 & \blue{GIFT-Eval-Pretrain}, \blue{Wiki Pageviews}, Google Trends, Synthetic. \\ \addlinespace

TiRex-1.1 \citep{auer2025tirex} & 35M & xLSTM & $2048^{(2)}$ & \blue{Subset of GIFT-Eval-Pretrain}, \blue{Subset of Chronos}, Synthetic. \\ \addlinespace

Chronos-2 \citep{ansari2025chronos2} & 120M & Encoder & 8192 & \red{Chronos}, High-volume Synthetic. \\

\bottomrule
\end{tabularx}
\end{table}

Table \ref{tab:tsfm-info} summarizes the key specifications of the baseline Time Series Foundation Models (TSFMs) evaluated in this study, including model size, backbone architecture, context length, and pretraining corpus composition. This information was collated from the original publications, official GitHub repositories, and Hugging Face model cards. We note the following specific configurations:
\begin{enumerate}
    \item \textbf{TabPFN:} While the model supports variable input lengths, we report a context length of 4096, consistent with the default setting in the official evaluation code.
    \item \textbf{TiRex:} We report the context length of 2048 based on the initial version described in \citet{auer2025tirex}. Although TiRex-1.1 introduces updates to address data leakage in the original release and some improvements, the context length is not explicitly redefined in the documentation; we therefore assume it remains unchanged.
    \item \textbf{Synthetic Data}: The inclusion of synthetic data has become a standard practice for pretraining TSFMs. Notably, many recent works generate their own synthetic corpora. While the generation methodologies are often briefly described in technical reports or publications, the exact synthetic datasets are rarely released.
\end{enumerate}

\newpage
\section{Statistical Analysis of Benchmark Results} \label{append:stats}

The GIFT-Eval and TIME protocols report aggregate scores over 97 and 98 test cases, respectively, where a case is a dataset $\times$ frequency $\times$ horizon combination. Since several of the gaps in Figure \ref{fig:gift-benchmark} and Table \ref{tab:time-benchmark} are small, we quantify the uncertainty of these aggregate comparisons across benchmark cases. All numbers below are computed from the case-level results of each model.

\subsection{Protocol} \label{append:stats-protocol}

\paragraph{Normalization and Aggregation} For case $i$, metric $m \in \{\text{MASE}, \text{CRPS}\}$, and model $M$, let $s^{(m)}_{M,i}$ denote the case-level score. Following the official protocols, each score is normalized by the same-case Seasonal-Naive score,
\begin{equation}
    x^{(m)}_{M,i} = \frac{s^{(m)}_{M,i}}{s^{(m)}_{\mathrm{SN},i}},
\end{equation}
and aggregated over a set of cases $I$ with the geometric mean, $A^{(m)}_{M}(I) = \exp\big(\frac{1}{|I|}\sum_{i \in I} \log x^{(m)}_{M,i}\big)$.

\paragraph{Paired Stratified Cluster Bootstrap} Cases from the same dataset--frequency pair share data and are not independent. We therefore treat each dataset--frequency pair as a cluster and keep all of its horizons together when resampling. Clusters with one horizon and clusters with three horizons are resampled separately (with replacement), so that each bootstrap replicate preserves the composition of the benchmark. GIFT-Eval has 34 one-horizon and 21 three-horizon clusters; the 74-case non-overlapping subset has 26 and 16; TIME has 26 and 24. For each replicate $b$ with resampled multiset of cases $I_b$, the same $I_b$ is used for both models of a comparison, and we compute the ratio of aggregate scores
\begin{equation}
    R^{(m,b)}_{A/B} = \frac{A^{(m)}_{A}(I_b)}{A^{(m)}_{B}(I_b)}.
\end{equation}
We draw 10{,}000 replicates (seed 20260728) and report the ratio on the full benchmark together with its 2.5th and 97.5th bootstrap percentiles as a 95\% confidence interval. A ratio below 1 favors the numerator model, and an interval that contains 1 indicates that the two models are not statistically distinguishable at this level.

\paragraph{Scope} The intervals are pointwise and are not adjusted for multiple comparisons. They capture variation across benchmark cases, not variation due to initialization, data ordering, or optimization: each full-scale variant corresponds to a single pretraining run, and we do not repeat pretraining across seeds.

\subsection{GIFT-Eval} \label{append:stats-gift}

Table \ref{tab:bootstrap-gift} compares the Zero-shot model with the three strongest baselines on the 97 GIFT-Eval cases. All six intervals include 1; the same analysis on the 74 non-overlapping cases leads to the same conclusion. The Zero-shot model is therefore competitive with, but not distinguishable from, these baselines. Case-level outputs of the Extended model on GIFT-Eval are not available, so it is not included here.

\begin{table}[h]
\centering
\caption{Paired cluster-bootstrap 95\% confidence intervals for the ratio of aggregate scores on \textbf{GIFT-Eval} (97 cases). A ratio below 1 favors the numerator model.}
\label{tab:bootstrap-gift}
\begin{tabular}{lcc}
\toprule
\textbf{Comparison} & \textbf{MASE ratio [95\% CI]} & \textbf{CRPS ratio [95\% CI]} \\
\midrule
Zero-shot / Chronos-2   & 1.013 [0.987, 1.041] & 0.995 [0.965, 1.022] \\
Zero-shot / TimesFM-2.5 & 1.003 [0.977, 1.032] & 0.985 [0.958, 1.012] \\
Zero-shot / Timer-S1    & 1.020 [0.982, 1.062] & 0.995 [0.963, 1.031] \\
\bottomrule
\end{tabular}
\end{table}

\subsection{TIME} \label{append:stats-time}

Table \ref{tab:bootstrap-time} reports the same analysis on the 98 TIME cases, where no case is seen by either of our models. The Extended model improves over the Zero-shot model on both metrics with intervals that exclude 1, and likewise over Timer-S1; it is on par with Chronos-2 on both metrics and with TimesFM-2.5 on MASE, while its CRPS interval against TimesFM-2.5 lies below 1. The Zero-shot model is on par with Chronos-2 and TimesFM-2.5 and improves over Timer-S1.

\begin{table}[h]
\centering
\caption{Paired cluster-bootstrap 95\% confidence intervals for the ratio of aggregate scores on \textbf{TIME} (98 cases). A ratio below 1 favors the numerator model.}
\label{tab:bootstrap-time}
\begin{tabular}{lcc}
\toprule
\textbf{Comparison} & \textbf{MASE ratio [95\% CI]} & \textbf{CRPS ratio [95\% CI]} \\
\midrule
Extended / Zero-shot    & 0.986 [0.975, 0.995] & 0.988 [0.977, 0.997] \\
Extended / Chronos-2    & 1.000 [0.986, 1.016] & 1.001 [0.986, 1.017] \\
Extended / TimesFM-2.5  & 0.990 [0.976, 1.005] & 0.981 [0.965, 0.998] \\
Extended / Timer-S1     & 0.943 [0.905, 0.975] & 0.946 [0.911, 0.975] \\
\midrule
Zero-shot / Chronos-2   & 1.014 [0.997, 1.032] & 1.013 [0.997, 1.030] \\
Zero-shot / TimesFM-2.5 & 1.004 [0.987, 1.021] & 0.994 [0.976, 1.012] \\
Zero-shot / Timer-S1    & 0.956 [0.916, 0.990] & 0.958 [0.923, 0.987] \\
\bottomrule
\end{tabular}
\end{table}

\subsection{Dispersion and Ranks across Cases} \label{append:stats-ranks}

Tables \ref{tab:ranks-gift} and \ref{tab:ranks-time} summarize how the normalized scores and the within-case ranks vary across cases. For each model, we report the sample standard deviation of $x^{(m)}_{M,i}$ over cases and the minimum, first quartile, median, third quartile, and maximum of its within-case rank. Ranks on GIFT-Eval are computed among the 14 models of Figure \ref{fig:gift-benchmark} with complete matched outputs, and ranks on TIME among the six models of Table \ref{tab:time-benchmark} for which case-level results are available; ties receive average ranks. On both benchmarks, every model ranks first on some cases and near the bottom on others, which reflects the heterogeneity of the benchmarks rather than a consistent ordering of the models.

\begin{table}[h]
\centering
\caption{Case-level dispersion on \textbf{GIFT-Eval} (97 cases). SD is the sample standard deviation of the Seasonal-Naive-normalized score across cases; ranks are min/Q25/median/Q75/max among 14 models.}
\label{tab:ranks-gift}
\begin{tabular}{lcccc}
\toprule
\textbf{Model} & \textbf{MASE SD} & \textbf{MASE ranks} & \textbf{CRPS SD} & \textbf{CRPS ranks} \\
\midrule
Zero-shot   & 0.202 & 1/2/4/7/14 & 0.198 & 1/2/5/7/14 \\
Chronos-2   & 0.179 & 1/2/3/6/14 & 0.192 & 1/2/4/6/14 \\
TimesFM-2.5 & 0.181 & 1/3/5/7/11 & 0.204 & 1/3/5/8/13 \\
Timer-S1    & 0.187 & 1/2/5/7/12 & 0.217 & 1/2/5/8/13 \\
\bottomrule
\end{tabular}
\end{table}

\begin{table}[h]
\centering
\caption{Case-level dispersion on \textbf{TIME} (98 cases). SD is the sample standard deviation of the Seasonal-Naive-normalized score across cases; ranks are min/Q25/median/Q75/max among six models.}
\label{tab:ranks-time}
\begin{tabular}{lcccc}
\toprule
\textbf{Model} & \textbf{MASE SD} & \textbf{MASE ranks} & \textbf{CRPS SD} & \textbf{CRPS ranks} \\
\midrule
Zero-shot   & 0.175 & 1/2/3/5/6    & 0.176 & 1/2/3/5/6 \\
Extended    & 0.159 & 1/1.25/3/4/6 & 0.170 & 1/2/3/4/6 \\
Chronos-2   & 0.150 & 1/1/2/4/6    & 0.163 & 1/1/3/4/6 \\
TimesFM-2.5 & 0.152 & 1/2/4/5/6    & 0.163 & 1/3/4/5.75/6 \\
Timer-S1    & 0.205 & 1/3/5/6/6    & 0.190 & 1/3/4/5/6 \\
\bottomrule
\end{tabular}
\end{table}

\newpage
\section{Additional Experimental Results and Visualizations}

\subsection{Additional Benchmarking Results}

\subsubsection{Pretraining Data Composition}

\begin{table}[htbp]
\centering
\caption{Ablation analysis of \textbf{pretraining data composition}. We compare models trained on individual data sources and composite datasets, as introduced in Section \ref{sec:exp-setup}. $^\dagger$The Extended corpus overlaps 23 GIFT-Eval test cases (Section \ref{sec:exp-setup}).}
\label{tab:dataset_ablation}

\begin{tabular}{lccc}
\toprule
\textbf{Pretrain Data} & \textbf{MASE} $\downarrow$ & \textbf{CRPS} $\downarrow$ & \textbf{Rank} $\downarrow$ \\
\midrule
KernelSynth & 0.832 & 0.584 & 12.220 \\
Real Only & 0.790 & 0.542 & 7.008 \\
Zero-shot & 0.712 & 0.483 & 4.267 \\
\rowcolor{gray!15} \textbf{Extended}$^\dagger$ & \textbf{0.696} & \textbf{0.480} & \textbf{3.383} \\
\bottomrule
\end{tabular}
\end{table}

\begin{figure}[h]
    \centering
    \includegraphics[width=\linewidth]{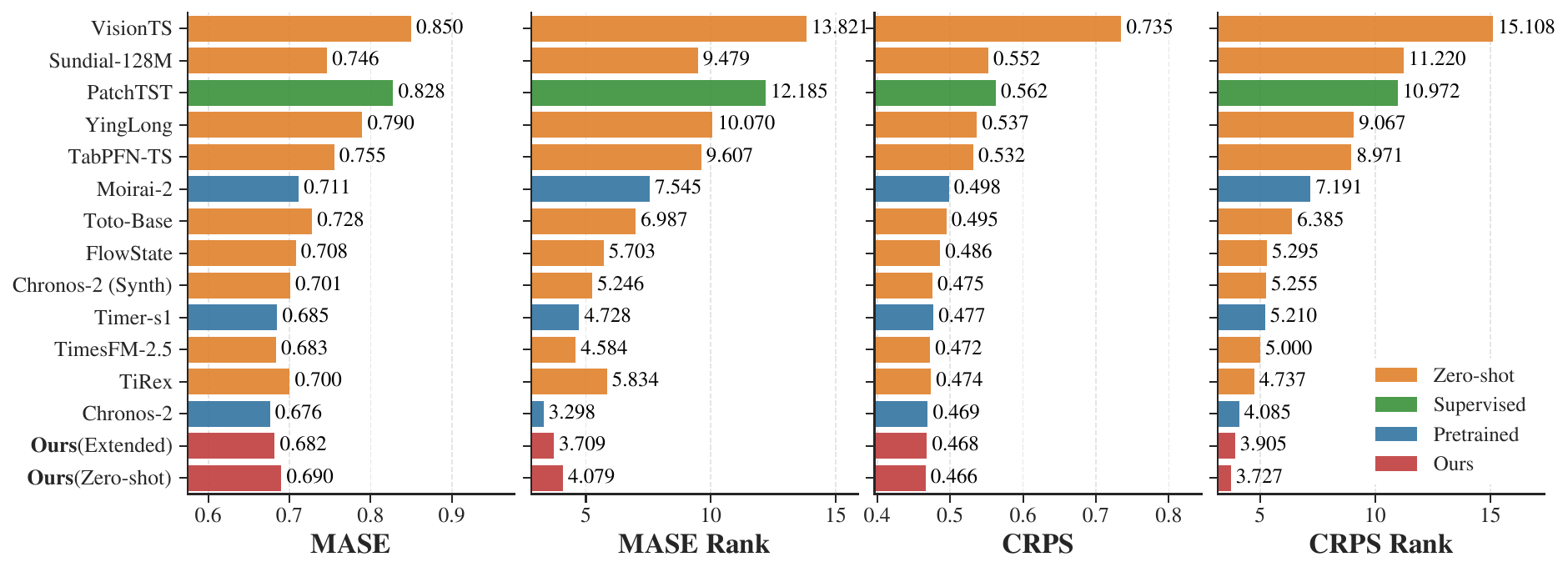}
    \caption{\textbf{Zero-shot} forecasting performance on the GIFT-Eval benchmark. The three metrics (MASE, CRPS, Rank) are aggregated with geometric mean over the \textbf{74 non-overlapping} test cases, none of which is seen by either of our models. Lower values indicate better performance. The methods are sorted based on Aggregated Rank.}
    \label{fig:gift-benchmark-no-leak}
\end{figure}

\subsubsection{Output Granularity}

\begin{table}[ht]
    \caption{Ablation study on \textbf{output probability granularity}. We compare the performance of models trained to predict 9, 21, and 99 quantiles.}
    \centering
    \label{tab:quantile_ablation}
    \begin{tabular}{lccc}
    \toprule
    Num. Quantiles & \textbf{MASE} $\downarrow$ & \textbf{CRPS} $\downarrow$ & \textbf{Rank} $\downarrow$ \\
    \midrule
    9  & 0.699 & 0.482 & 3.758 \\
    21 & 0.700 & 0.481 & 4.087 \\
    \rowcolor{gray!15} \textbf{99} & \textbf{0.696} & \textbf{0.480} & \textbf{3.598} \\
    \bottomrule
    \end{tabular}
\end{table}

We investigate the impact of the output distribution's resolution by varying the number of predicted quantiles $K=\{9, 21, 99\}$ with the corresponding quantile levels:
\begin{itemize}[nosep, leftmargin=0.3cm]
    \item $K=9$: $Q=\{0.1, 0.2, \dots, 0.9\}$,
    \item $K=21$: $Q=\{0.01, 0.05, 0.1, \dots, 0.9, 0.95, 0.99\}$,
    \item $K=99$: $Q=\{0.01, 0.02, \dots, 0.99\}$.
\end{itemize}
As shown in Table \ref{tab:quantile_ablation}, the three variants result in very similar performance, while $K=99$ has slight advantage over the coarser predictions. This indicates that while a coarse set of quantiles is sufficient to capture the general distributional, higher granularity is more general and precise for fine-grained predictions. Note that, for fair comparison, the CRPS score of the three variants are computed only on the quantile levels $\{0.1, 0.2, \dots, 0.9\}$.

\subsubsection{Patch Size} \label{append:patch-size}

\begin{table}[ht]
    \centering
    \caption{Ablation study on \textbf{patch size} $L$ on GIFT-Eval. Tokens is the number of patches at the default context length $T=8192$. The three models form a separate set of pretraining runs; see the text for the training setup.}
    \label{tab:patch_ablation}
    \begin{tabular}{lcccc}
    \toprule
    \textbf{Patch Size} & \textbf{Tokens} & \textbf{MASE} $\downarrow$ & \textbf{CRPS} $\downarrow$ & \textbf{Rank} $\downarrow$ \\
    \midrule
    8  & 1024 & \textbf{0.714} & \textbf{0.487} & 5.495 \\
    \rowcolor{gray!15} \textbf{16} & 512 & 0.719 & \textbf{0.487} & \textbf{5.340} \\
    32 & 256 & 0.724 & 0.497 & 6.722 \\
    \bottomrule
    \end{tabular}
\end{table}

\begin{table}[ht]
    \centering
    \caption{Patch-size ablation broken down by sampling frequency and forecast term. Each entry is the Seasonal-Naive-normalized geometric mean over the cases in the group. Minute-level groups the 5-, 10-, and 15-minute datasets; Monthly+ groups the monthly, quarterly, and yearly datasets.}
    \label{tab:patch_ablation_breakdown}
    \begin{tabular}{lccccccc}
    \toprule
    & & \multicolumn{3}{c}{\textbf{MASE} $\downarrow$} & \multicolumn{3}{c}{\textbf{CRPS} $\downarrow$} \\
    \cmidrule(lr){3-5} \cmidrule(lr){6-8}
    \textbf{Group} & \textbf{Cases} & $L=8$ & $L=16$ & $L=32$ & $L=8$ & $L=16$ & $L=32$ \\
    \midrule
    10-second    & 6  & \textbf{0.676} & 0.705 & 0.715 & \textbf{0.478} & 0.546 & 0.514 \\
    Minute-level & 30 & 0.734 & \textbf{0.723} & 0.737 & 0.500 & \textbf{0.489} & 0.511 \\
    Hourly       & 31 & \textbf{0.693} & 0.696 & 0.697 & 0.434 & \textbf{0.432} & 0.439 \\
    Daily        & 15 & 0.661 & \textbf{0.660} & 0.671 & 0.416 & \textbf{0.411} & 0.418 \\
    Weekly       & 8  & 0.779 & \textbf{0.773} & 0.798 & 0.624 & \textbf{0.621} & 0.631 \\
    Monthly+     & 7  & \textbf{0.823} & 0.906 & 0.843 & \textbf{0.780} & 0.809 & 0.811 \\
    \midrule
    Short        & 55 & \textbf{0.687} & 0.699 & 0.701 & \textbf{0.508} & 0.509 & 0.516 \\
    Medium       & 21 & 0.736 & \textbf{0.735} & 0.741 & \textbf{0.458} & 0.467 & 0.475 \\
    Long         & 21 & 0.770 & \textbf{0.756} & 0.770 & 0.464 & \textbf{0.455} & 0.469 \\
    \bottomrule
    \end{tabular}
\end{table}

We vary the patch size $L \in \{8, 16, 32\}$ while keeping the context length, data recipe, and all other hyperparameters fixed, so that the number of tokens per sample changes from 1024 to 256. This ablation was conducted after the main experiments on a different cluster (32 NVIDIA A100-80G GPUs instead of 96 V100-32G GPUs; see Appendix \ref{append:framework}) under the Zero-shot data recipe, so the three runs are compared with each other rather than with the ablations above; the $L=16$ run therefore differs slightly from the Zero-shot model in Figure \ref{fig:gift-benchmark}. Table \ref{tab:patch_ablation} reports the GIFT-Eval aggregates. $L=8$ and $L=16$ obtain the same CRPS and differ by 0.005 in MASE; across the 97 cases, each is better than the other on roughly half of them (49 vs. 48 cases on CRPS), and 74 cases differ by less than 5\%. $L=32$ is consistently worse: its CRPS is 2\% higher, and it loses to $L=16$ on 65 of the 97 cases.

Table \ref{tab:patch_ablation_breakdown} breaks the comparison down by sampling frequency and forecast term. On the 84 cases sampled at minute to weekly resolution, $L=8$ and $L=16$ differ by only 1--2\%, $L=32$ is within 2\% of $L=16$ except on minute-level data (4.5\% in CRPS), and $L=16$ is the best or on par with the best. The advantage of $L=8$ is concentrated in the six 10-second cases, where it improves CRPS over $L=16$ by 24\% at the medium and 10\% at the long horizon, and in short-horizon forecasts, where it is slightly better; at long horizons on minute-level data it is about 6\% worse than $L=16$. The degradation of $L=32$ is uniform across terms (about 1.4\% in median CRPS) and most visible on minute-level data. The Monthly+ group contains only seven cases and is dominated by two of them (\texttt{car\_parts} and \texttt{us\_births}), so we do not draw conclusions from it. Overall, since $L=8$ doubles the token count without an aggregate gain, the choice $L=16$ is still a reasonable default.

\subsubsection{Scaling Training Budget}

\begin{figure}[h]
    \centering
    \includegraphics[width=0.75\linewidth]{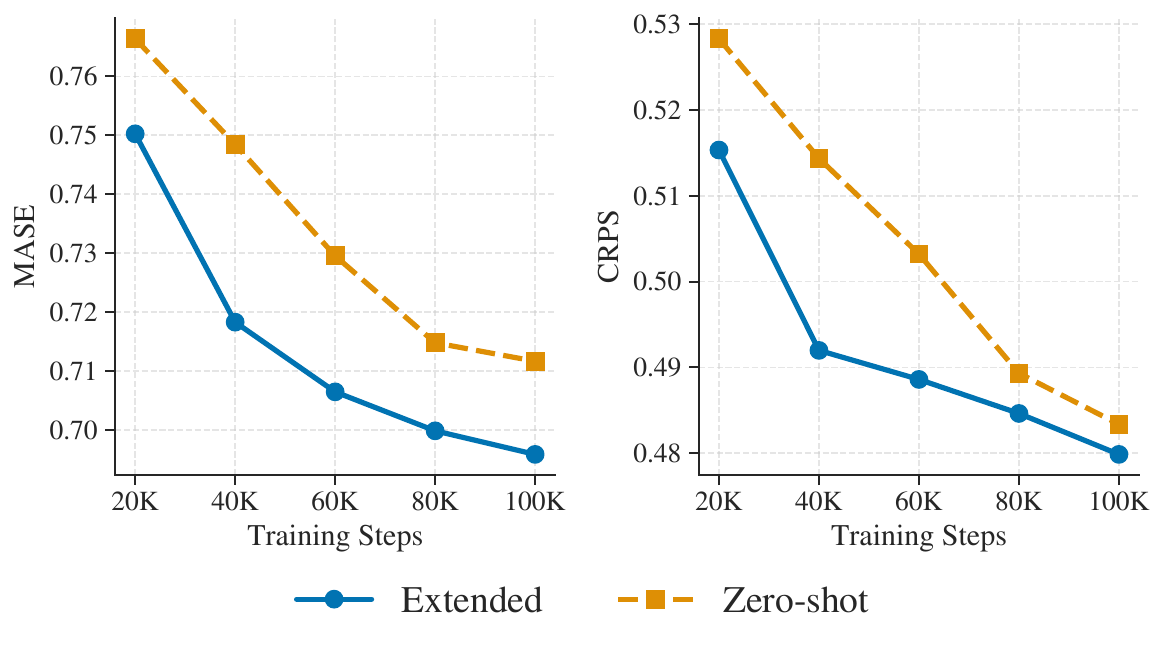}
    \caption{\textbf{Impact of training scaling on forecasting performance}: We evaluate the intermediate model checkpoints during training. MASE and CRPS for both the Extended and Zero-shot variants exhibit a consistent monotonic improvement.}
    \label{fig:train-steps}
\end{figure}

We investigate the relationship between training duration and downstream performance by evaluating checkpoints throughout the pretraining process (see Figure \ref{fig:train-steps}). The results show a steady improvement in both MASE and CRPS. The trajectories suggest that the models have not yet reached their asymptotic limit, implying that further performance gains can be achieved simply by extending the training budget.

\subsubsection{Details of the Scaling Study} \label{append:scaling-details}

Figure \ref{fig:model-scaling} reports two one-factor sweeps for each pretraining corpus (Zero-shot and Extended). The width sweep fixes $N_{\text{layer}}=20$ and varies $d \in \{384, 512, 768, 1024\}$; the depth sweep fixes $d=1024$ and varies $N_{\text{layer}} \in \{4, 8, 12, 20\}$. All runs use the same data recipe, 100k training steps, context length 8192, and effective batch size; they are therefore matched in training steps and nominal token budget but not in FLOPs, and each point is a single pretraining run. The sweeps characterize the accuracy--size trade-off within the Transformer backbone and should not be read as a universal scaling law.

\subsubsection{Summary of Design Choices} \label{append:design-summary}

We summarize the empirical evidence behind the default configuration in Table \ref{tab:hyperparameters}.
\begin{itemize}[nosep, leftmargin=0.3cm]
    \item \textbf{Model size} (Figure \ref{fig:model-scaling}): at similar parameter counts, deeper and narrower variants perform better, so parameter allocation matters beyond the total size.
    \item \textbf{Context length} (Table \ref{tab:context_ablation}): increasing $T$ from 2048 to 4096 to 8192 lowers CRPS from 0.500 to 0.497 to 0.480.
    \item \textbf{CPM block size} (Table \ref{tab:cpm_ablation}): $N_{\text{CPM}} = 1/4/8$ gives CRPS 0.514/0.482/0.480; most of the gain is obtained by $N_{\text{CPM}}=4$.
    \item \textbf{Masking ratio} (Table \ref{tab:mask_ratio_ablation}): 40\% gives the best MASE and CRPS, while 20\% gives the best rank; the preferred ratio depends on the metric.
    \item \textbf{Number of quantiles} (Table \ref{tab:quantile_ablation}): 9/21/99 quantiles give CRPS 0.482/0.481/0.480; the gain from denser quantiles is small.
    \item \textbf{Training steps} (Figure \ref{fig:train-steps}): both metrics improve monotonically from 20k to 100k steps within a single run.
    \item \textbf{Patch size} (Table \ref{tab:patch_ablation}, separate runs): $L=8$ and $L=16$ give the same CRPS (0.487) and $L=32$ is worse (0.497); $L=16$ is a reasonable default choice since $L=8$ doubles the token count without an aggregate gain.
\end{itemize}


\newpage
\subsection{When will it fail? Explore the maximum supported forecast length}
\begin{figure}[h]
    \centering
    \includegraphics[width=0.9\linewidth]{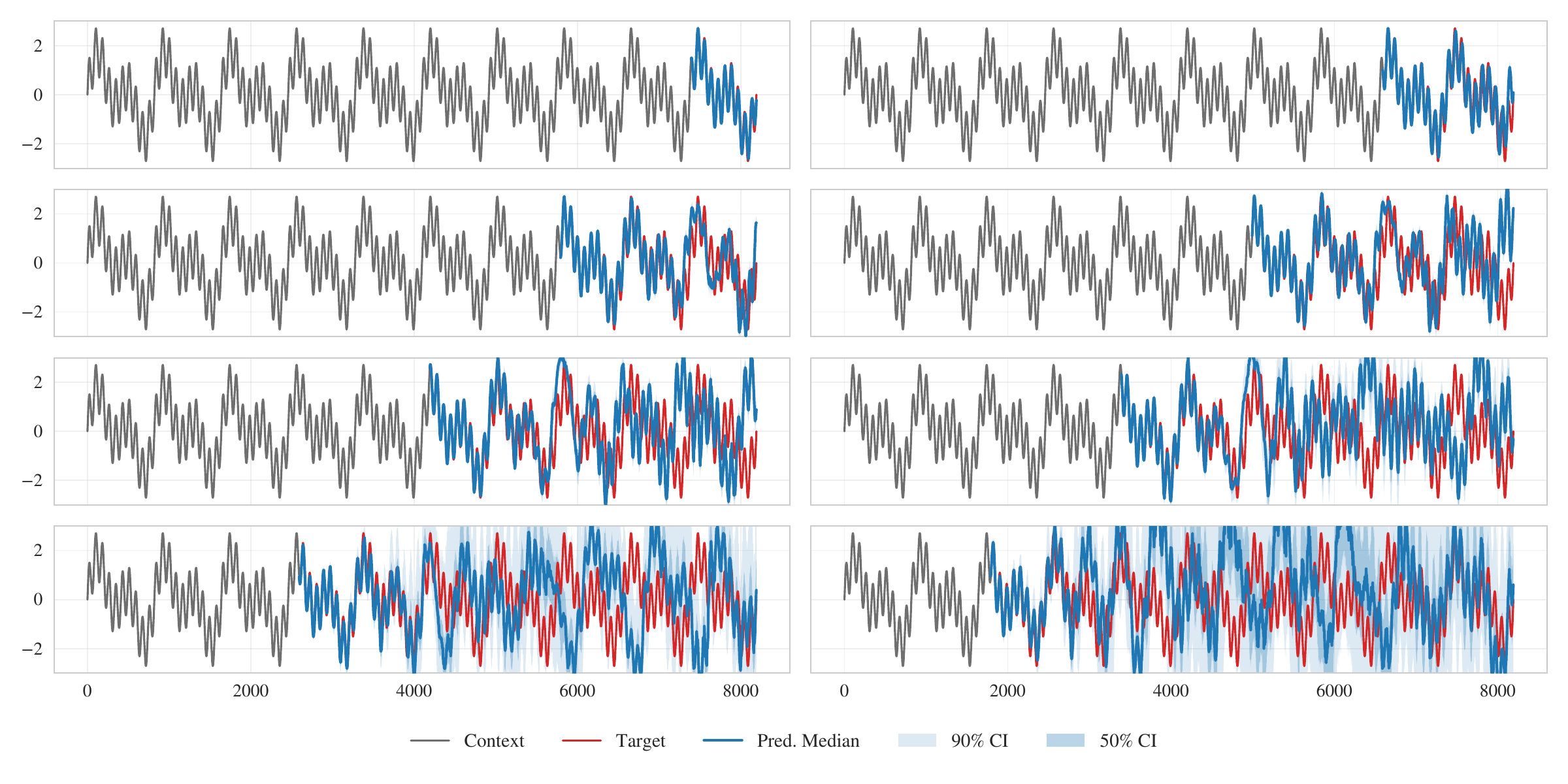}
    \caption{Examples of predictions on a synthetic signal with different predictions lengths.}
    \label{fig:synth_pred_len}
\end{figure}

Our generalized pretraining masking strategy offers the distinct advantage of flexible forecast lengths. In Figure \ref{fig:synth_pred_len}, we investigate the effective maximum forecasting horizon supported by the model. We evaluate performance on a synthetic multi-frequency sine wave across forecast lengths $\{800, 1600, \dots, 6400\}$. We observe that while predictions remain accurate for shorter horizons $\{800, 1600\}$, performance gradually degrades as the forecast length increases. These findings imply the following:
\begin{itemize}[nosep, leftmargin=0.3cm]
    \item For optimal performance, the maximum reliable forecast length is approximately 2000 timestamps. Extending beyond this limit can be achieved with autoregressive decoding.
    \item Neural models may struggle to extrapolate "simple" periodic cases over long horizons, whereas such patterns are often trivially predicted by classical statistical models.
\end{itemize}

\newpage
\subsection{Horizon Dependence on Real Data} \label{append:horizon}

To complement the synthetic stress test above, Table \ref{tab:horizon} reports absolute case-level MASE and CRPS on three TIME datasets at their short, medium, and long horizons, for our two models and the three strongest baselines. The three datasets were chosen to span different sampling frequencies (hourly and 20-minute), horizon lengths (9 to 336 steps), and domains. The effect of the horizon depends on both the dataset and the metric: CRPS decreases with the horizon on \texttt{Australia\_Solar/H} for all five models, whereas both errors increase with the horizon on \texttt{Coastal\_T\_S/20T} and \texttt{CPHL/H}. The ordering of the models also changes across datasets and horizons. We therefore do not draw general conclusions about horizon dependence from these three examples.

\begin{table}[h]
\centering
\small
\caption{Case-level MASE and CRPS on three TIME datasets at short (S), medium (M), and long (L) horizons. Values are absolute (not normalized) case-level scores; lower is better.}
\label{tab:horizon}
\begin{tabular}{llcccccc}
\toprule
\textbf{Dataset (S/M/L)} & \textbf{Model} & \textbf{MASE-S} & \textbf{MASE-M} & \textbf{MASE-L} & \textbf{CRPS-S} & \textbf{CRPS-M} & \textbf{CRPS-L} \\
\midrule
\multirow{5}{*}{\begin{tabular}[c]{@{}l@{}}Australia\_Solar/H\\(24/72/168)\end{tabular}}
 & Extended    & 0.5746 & 0.6031 & 0.5789 & 0.2320 & 0.2102 & 0.1972 \\
 & Zero-shot   & 0.6278 & 0.6745 & 0.6832 & 0.2322 & 0.2215 & 0.2193 \\
 & Chronos-2   & 0.5689 & 0.6116 & 0.6182 & 0.2583 & 0.2250 & 0.2103 \\
 & TimesFM-2.5 & 0.6053 & 0.6114 & 0.6012 & 0.2657 & 0.2237 & 0.2067 \\
 & Timer-S1    & 0.5861 & 0.5986 & 0.5776 & 0.2519 & 0.2215 & 0.2038 \\
\midrule
\multirow{5}{*}{\begin{tabular}[c]{@{}l@{}}Coastal\_T\_S/20T\\(9/72/216)\end{tabular}}
 & Extended    & 0.4701 & 0.9593 & 1.5430 & 0.0018 & 0.0040 & 0.0062 \\
 & Zero-shot   & 0.4421 & 0.9737 & 1.5763 & 0.0017 & 0.0042 & 0.0067 \\
 & Chronos-2   & 0.4837 & 0.9894 & 1.4011 & 0.0019 & 0.0041 & 0.0055 \\
 & TimesFM-2.5 & 0.4038 & 0.9137 & 1.3247 & 0.0015 & 0.0038 & 0.0055 \\
 & Timer-S1    & 0.5812 & 1.0121 & 1.5084 & 0.0022 & 0.0041 & 0.0060 \\
\midrule
\multirow{5}{*}{\begin{tabular}[c]{@{}l@{}}CPHL/H\\(24/168/336)\end{tabular}}
 & Extended    & 1.2066 & 1.7346 & 2.0068 & 0.2201 & 0.3007 & 0.3402 \\
 & Zero-shot   & 1.2206 & 1.7413 & 1.9741 & 0.2214 & 0.2977 & 0.3302 \\
 & Chronos-2   & 1.1943 & 1.7754 & 2.0681 & 0.2199 & 0.3050 & 0.3557 \\
 & TimesFM-2.5 & 1.2108 & 1.7389 & 1.9857 & 0.2182 & 0.2931 & 0.3347 \\
 & Timer-S1    & 1.2209 & 1.7995 & 2.0225 & 0.2226 & 0.3141 & 0.3645 \\
\bottomrule
\end{tabular}
\end{table}

\newpage

\subsection{Visualizations of Predictions}
\begin{figure}[h]
    \centering
    \includegraphics[width=\linewidth]{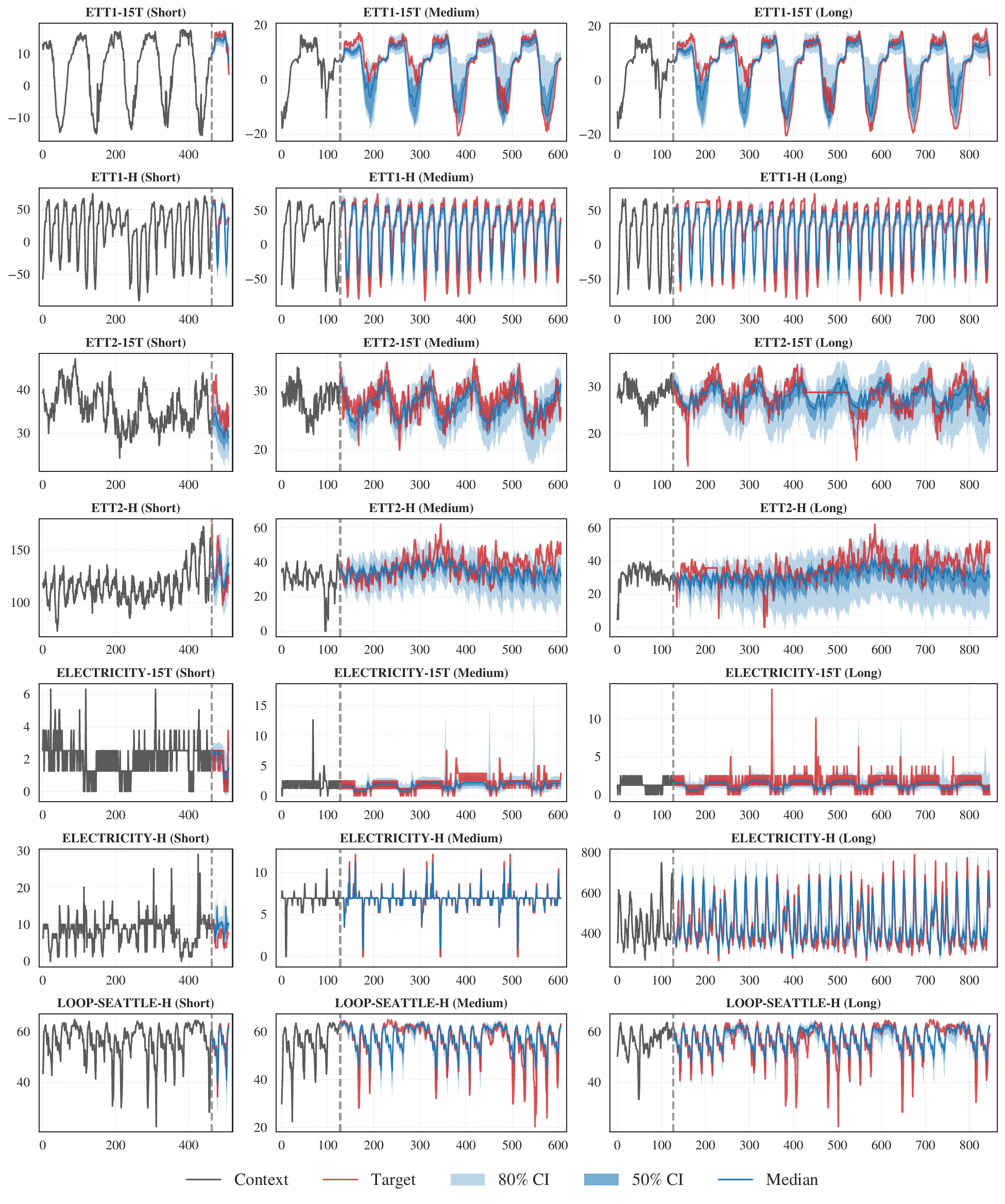}
    \caption{Examples of predictions from our \textbf{Extended} model on datasets with significant seasonal patterns.}
    \label{fig:pretrained-seaonal}
\end{figure}

\newpage

\begin{figure}[h]
    \centering
    \includegraphics[width=\linewidth]{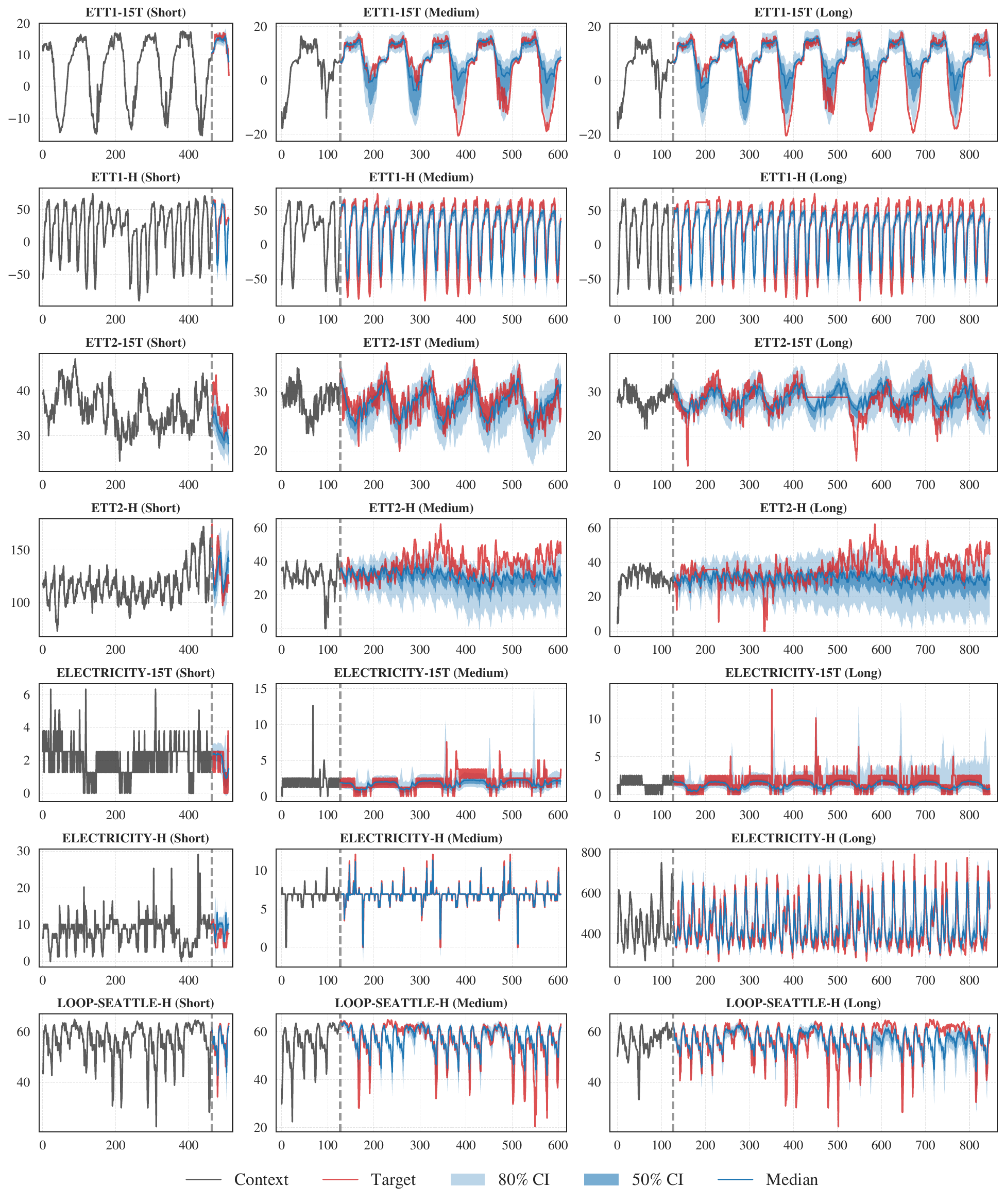}
    \caption{Examples of predictions from our \textbf{Zero-shot} model on datasets with significant seasonal patterns.}
    \label{fig:zero-shot-seasonal}
\end{figure}

\newpage

\begin{figure}[h]
    \centering
    \includegraphics[width=0.95\linewidth]{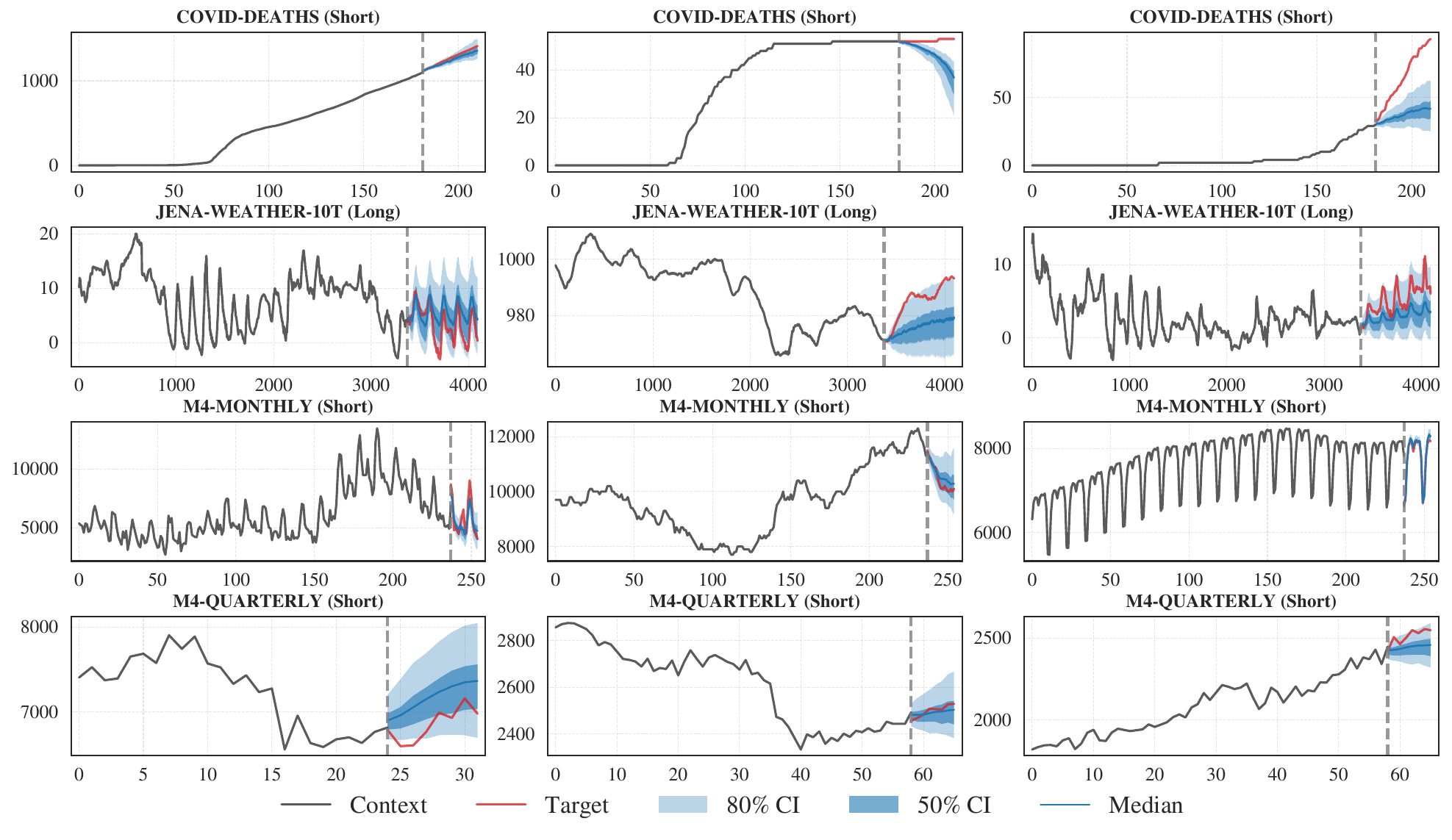}
    \caption{Examples of predictions from our \textbf{Extended} model on datasets with significant trend patterns.}
    \label{fig:pretrained-trend}
\end{figure}

\begin{figure}[h]
    \centering
    \includegraphics[width=0.95\linewidth]{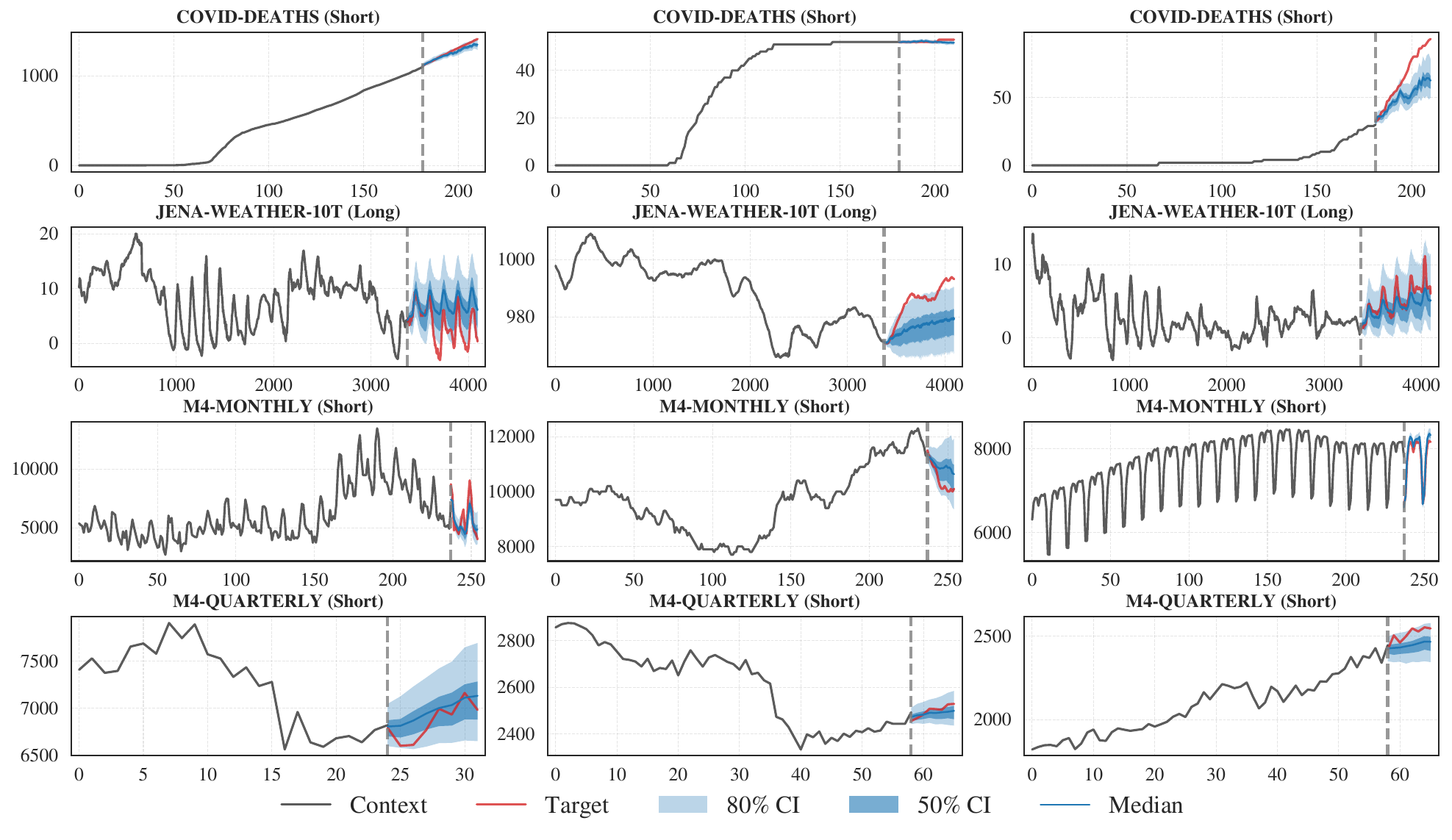}
    \caption{Examples of predictions from our \textbf{Zero-shot} model on datasets with significant trend patterns.}
    \label{fig:zero-shot-trend}
\end{figure}

\newpage

\begin{figure}[h]
    \centering
    \includegraphics[width=0.95\linewidth]{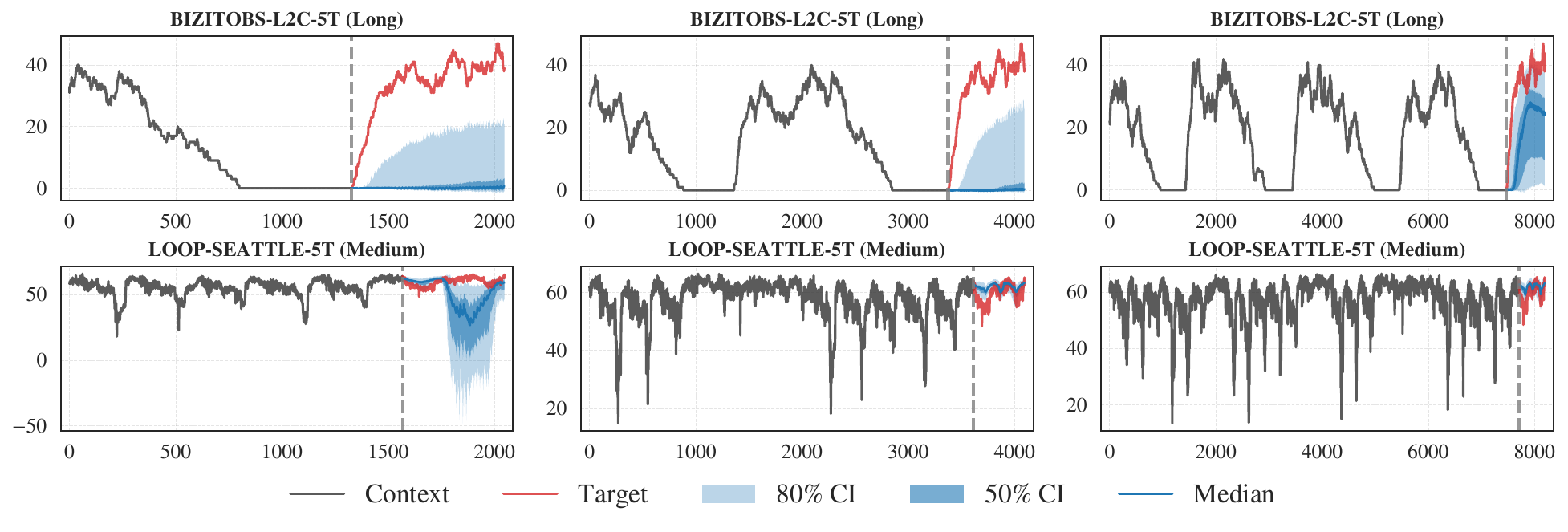}
    \caption{Example of the predictions from models with context length (left) 2048, (middle) 4096, and (right) 8192. Shorter context lengths may fail to capture periodicity and produce poor forecasts, while longer context lengths are more robust for general usage.}
    \label{fig:context-compare}
\end{figure}

\begin{figure}[h]
    \centering
    \includegraphics[width=0.95\linewidth]{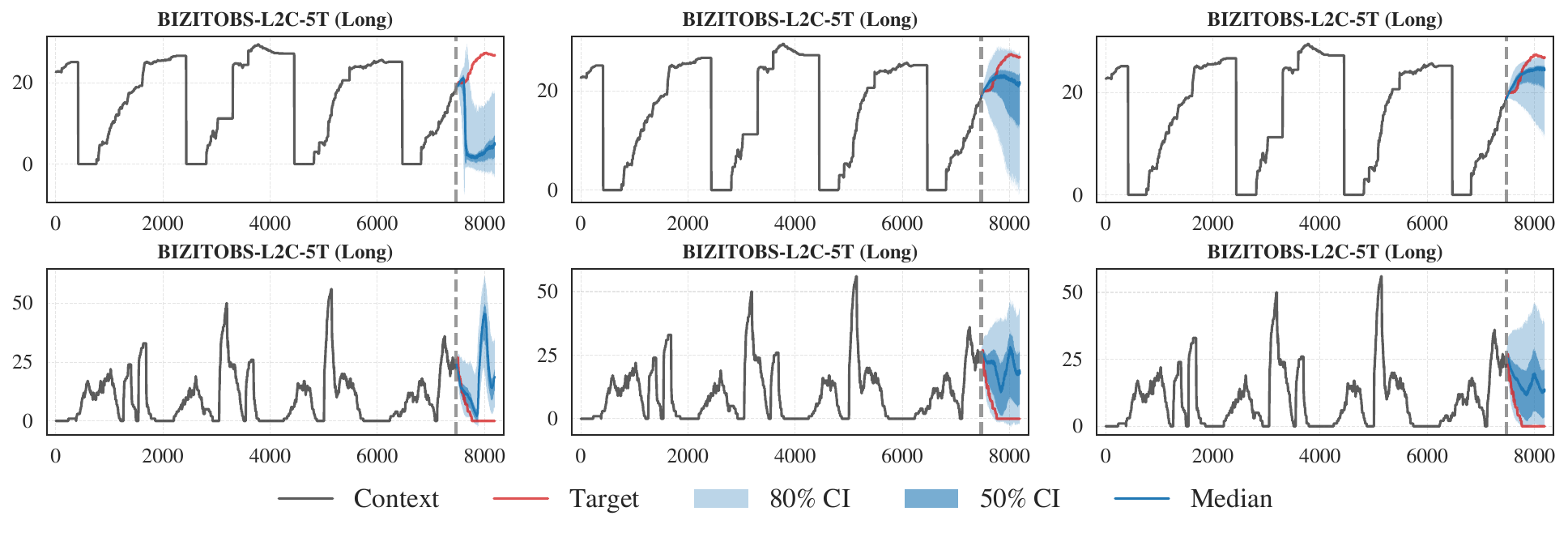}
    \caption{Examples of the predictions from models pretrained with CPM Block Size (left) $N_{\text{CPM}}=1$, (middle) $N_{\text{CPM}}=4$, and (right) $N_{\text{CPM}}=8$. Larger $N_{\text{CPM}}$ encourages the model to infer the underlying time series dynamics, providing better forecasts for general usage.}
    \label{fig:cpm-compare}
\end{figure}

\newpage
\subsection{Inference Time} \label{append:speed}

\begin{figure}[h]
    \centering
    \includegraphics[width=0.7\linewidth]{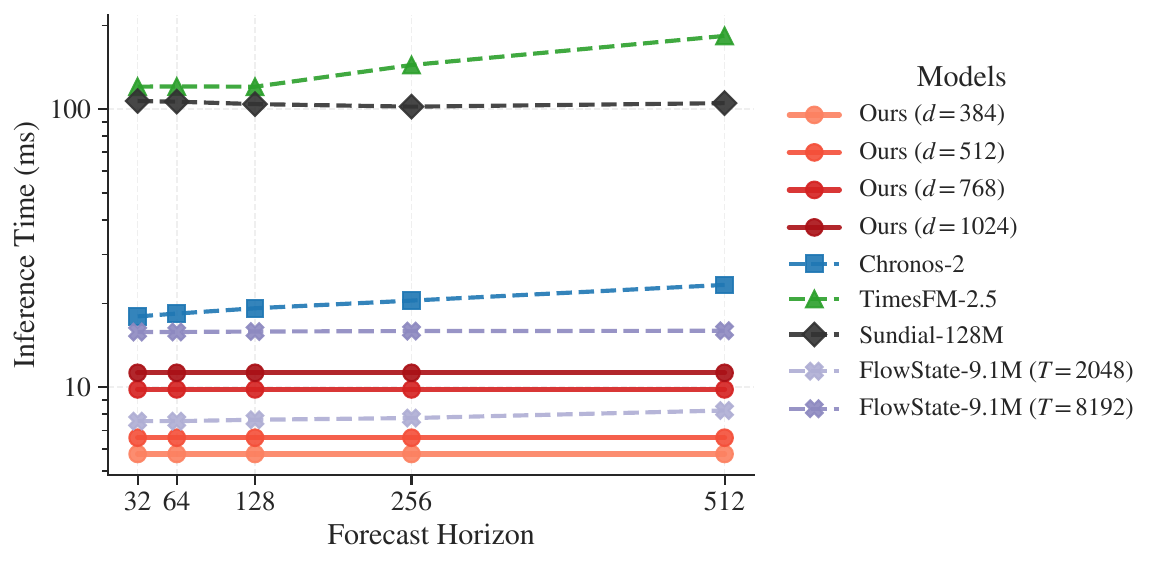}
    \caption{Inference time of our models and other TSFMs. Benefiting from high mask ratio, single-step inference, and architecture simplicity, our models also excel in inference time which is invariant to the forecast horizon.}
    \label{fig:inference-time}
\end{figure}
Figure \ref{fig:inference-time} benchmarks the inference latency of our proposed architecture against state-of-the-art TSFMs. All evaluations were conducted on an NVIDIA RTX 6000 Blackwell GPU using a batch size of 1. To ensure a rigorous comparison, we set the input context length to $T=8192$ for all compatible models, while adjusting for architectural constraints in baselines such as Sundial ($T=2880$). While FlowState is pretrained with $T=2048$, it also supports longer context length as a state-space model. Therefore, we include the evaluation for both cases and compare their inference speed. Results for comparison models were derived using their official implementations for inference. We note that TiRex is excluded from this specific comparison due to hardware-specific compilation incompatibilities with its custom state-space model kernels.

\end{document}